\documentclass[a4paper,fleqn]{cas-dc}

\usepackage[numbers]{natbib}

\usepackage{graphicx}
\usepackage{textcomp}
\usepackage{multirow}
\usepackage{colortbl}
\usepackage{arydshln}
\usepackage{hyperref}
\usepackage{makecell}
\usepackage{pifont}
\usepackage{booktabs}

\newcommand{\xmark}{\ding{55}}%

\begin{document}
\let\WriteBookmarks\relax
\def\floatpagepagefraction{1}
\def\textpagefraction{.001}
\shorttitle{MICRO: A Surgery-Phase Retinal Image Rigid Registration Benchmark}
\shortauthors{M. Gong et~al.}

\title [mode = title]{MICRO: A Surgery-Phase Retinal Image Rigid Registration Benchmark and One-Step Frequency-Structure Similarity Registration Learning}



\author[1]{Mingdao Gong}[style=chinese]
\fnmark[1] 

\author[2]{Yan Hu}
\fnmark[1] 

\author[3]{Mingzhen Yuan}

\author[1]{Jialu Zhang}

\author[1]{Junming Feng}

\author[1]{Yiming Jiang}

\author[3]{Jing Ma}

\author[3]{Jinghua Liu}

\author[4]{Huazhu Fu}

\author[3]{Hai Lu}
\cormark[1]
\ead{trdr_luhai2017@163.com}

\author[5]{Xiaoqing Zhang}
\cormark[1]
\ead{11930927@mail.sustech.edu.cn}

\author[1]{Jiang Liu}
\cormark[2] 
\ead{liuj@sustech.edu.cn}

\cortext[cor1]{Corresponding author}
\cortext[cor2]{Principal corresponding author}
\fntext[fn1]{Mingdao Gong and Yan Hu contributed equally to this work.}

\credit{Conceptualization of this study, Methodology, Software}


\affiliation[1]{organization={Research Institute of Trustworthy Autonomous Systems and Department of Computer Science and Engineering, Southern University of Science and Technology},
                city={Shenzhen},
                country={China}}

\affiliation[2]{organization={Institute of AI for Industries, Chinese Academy of Sciences},
                country={China}}

\affiliation[3]{organization={Beijing Tongren Eye Center, Beijing Tongren Hospital, Capital Medical University},
                addressline={Beijing Ophthalmology and Visual Sciences Key Laboratory},
                city={Beijing},
                country={China}}

\affiliation[4]{organization={Institute of High Performance Computing (IHPC), Agency for Science, Technology and Research (A*STAR)},
                country={Singapore}}

\affiliation[5]{organization={Center for High Performance Computing and Shenzhen Key Laboratory of Intelligent Bioinformatics, Shenzhen Institutes of Advanced Technology, Chinese Academy of Sciences},
                city={Shenzhen},
                postcode={518055},
                country={China}}

\nonumnote{All clinical data collection in this study was approved by the Ethics Committee of Beijing Tongren Hospital (Approval No.: TREC2022-XJS05). Written informed consent was obtained from all participants in strict adherence to the tenets of the Declaration of Helsinki.}

\begin{abstract}
Precise multimodal retinal image rigid registration (MRIRG) is essential for aligning cross-modal vessel structures, offering critical guidance for ophthalmic surgery. Despite its significance, the advancement of MRIRG is hindered by the lack of publicly available multimodal retinal image rigid registration datasets spanning various ophthalmic surgery phases (e.g., preoperative, intraoperative, and postoperative). To address this issue, we introduce the MICRO dataset, a large-scale MRIRG benchmark, comprising 1,365 cross-modal retinal image pairs with standardized annotations across all ophthalmic surgery phases. Moreover, existing deep registration methods achieve competitive MRIRG results via a two-step segmentation-then-detection framework, but segmentation quality heavily constrains their performance. We argue that this dependency stems from an implicit need for keypoint topology constraints, which have not been exploited for MRIRG through a one-step registration framework. Inspired by the above analysis, we propose a novel One-step Frequency-Structure Similarity Registration ($\text{OFS}^{2}\text{Reg}$) learning framework for MRIRG that leverages the potential of cross-modal vessel structure similarities to construct keypoint topology constraints under a shared detection network. To be specific, $\text{OFS}^{2}\text{Reg}$ applies a well-designed frequency-guided structure preservation descriptor to construct coarse keypoint topology constraints in descriptors by highlighting significant vessel structures. Additionally, we develop a new cross-modal vessel structure similarity learning method that refines cross-modal keypoint topology constraints to precisely align pixel-level cross-modal vessel structures by leveraging cross-modal vessel structure similarity priors. 
Extensive experiments on the MICRO dataset demonstrate that our $\text{OFS}^{2}\text{Reg}$ generally achieves state-of-the-art performance across different surgical-phase subsets. For example, on the MICRO-pre2post subset, $\text{OFS}^{2}\text{Reg}$ reduces RMSE by $1.02$ pixels and improves mAUC by $0.133$ compared with the one-step SuperRetina.

\end{abstract}


\maketitle

\section{Introduction}
\label{sec:intro}

\begin{figure}
    \centering
    \includegraphics[width=1\linewidth]{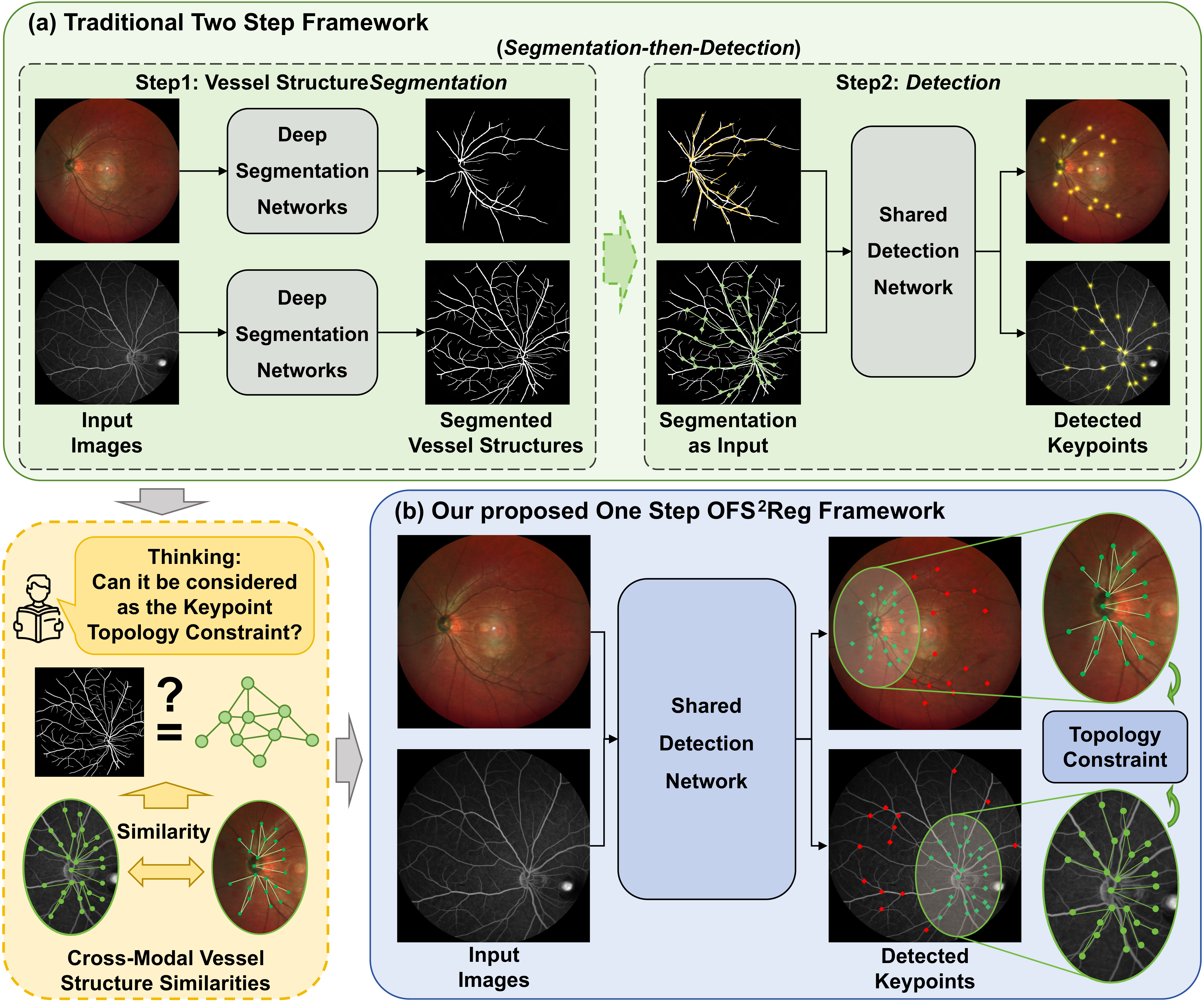}
    \caption{(a) Limitations of the two-step registration framework: Current deep registration methods typically conduct multimodal retinal image rigid registration (MRIRG) through a segmentation-then-detection framework. However, their performance is heavily contingent on the accuracy of vessel structure segmentation step across different retinal image modalities. We suggest that the segmentation step can be viewed as the keypoint topology constraint construction—a critical yet previously overlooked aspect that guides the detection network to align cross-modal vessel structures accurately. (b) Our proposed one-step registration framework: One-step Frequency-Structure Similarity Registration learning framework conducts multimodal retinal image rigid registration by fully leveraging the potential of cross-modal vessel structure similarities to construct keypoint topology constraints based on the single shared detection network.}
    \label{fig:problem}
\end{figure}

Retinal images acquired from multimodal retinal imaging systems capture diverse retinal structures of the eye, such as color fundus (CF), fluorescein angiography (FA), and ultra-widefield (UWF) imaging~\cite{10944285}. Each retinal image modality has a unique impact on various ophthalmic surgery phases for corresponding ocular diseases (e.g., pathological myopia), encompassing preoperative planning, intraoperative guidance, and postoperative quality assessment. Moreover, multimodal retinal image rigid registration (\textbf{MRIRG}) enables the comparison and alignment of vessel structures across different retinal image modalities, which significantly affects different ophthalmic surgery phases and has attracted growing attention from diverse perspectives~\cite{nie2024medical}. Similar registration and tracking techniques have also been explored in other image-guided surgical scenarios to support intraoperative navigation and dynamic anatomical alignment~\cite{cao2026robot,geng2025double}.

Over the years, scholars have released several retinal image registration datasets to promote the development of retinal image registration and corresponding techniques~\cite{hu2025coph100,nie2025reparameterized}, including single-modality datasets (e.g., FLoRI21~\cite{ding2021flori21}) and dual-modality datasets (e.g., MEMO~\cite{wang2024memo}). 
However, these datasets typically focus on fixed modality combinations, and few cover the variety of modality pairs encountered across surgical phases.
That is, they lack cross-modal retinal image pairs spanning all phases of ophthalmic surgery. To bridge this gap, this paper introduces a large-scale \textbf{M}ultimodal retinal \textbf{I}mage rigid registration benchmark for \textbf{C}linical \textbf{R}etinal \textbf{O}perations, named \textbf{MICRO}, comprising 1,365 cross-modal retinal image pairs from different ophthalmic surgery phases, with standardized annotations provided by four experienced clinicians. MICRO will be released as a publicly available benchmark for the comprehensive evaluation of MRIRG methods, advancing the development of the MRIRG field.

With the advent of deep learning techniques, deep registration methods have been increasingly applied to MRIRG and have achieved good performance. Wang et al.~\cite{wang2021robust} proposed a multimodal content-adaptive retinal image registration method to align vessel structures from cross-modal retinal images. They first applied a deep segmentation network to segment vessel structures of different retinal image modalities, which are then used as inputs for detection networks to detect informative keypoints, aiming to match cross-modal vessel structures accurately. 
Similarly, An~et al.~\cite{an2022self} developed a self-supervised rigid registration framework for cross-modal vessel structure alignment, consisting of a deep vessel structure segmentation step and a detection step. 
Veturi~\textit{et al.}~\cite{veturi2024eyeliner} presented the EyeLiner pipeline, which isolates peripheral vessel structures by masking unstable optic disk regions through a segmentation network, and subsequently performs keypoint detection on the remaining vessel structures.
Here, we conclude that current deep registration methods for MRIRG follow the segmentation-then-detection framework, as illustrated in Fig.~\ref{fig:problem}(a). In the segmentation step, existing methods typically employ deep segmentation networks to obtain vessel structures that contain keypoint topology constraints. Followed by the detection step, they take the segmented vessel structures as the inputs that allow the detection network to detect significant keypoints for aligning cross-modal vessel structures accurately. Moreover, we suggest that the segmentation step in the segmentation-then-detection framework can be considered as the keypoint topology constraint construction procedure, offering a unique perspective to rethink deep registration network design for MRIRG.  
Here, we define the concept of keypoint topology constraints in MRIRG for the segmentation-then-detection framework, which utilizes obtained vessel structures based on the segmentation step to construct inherent topology relationships among keypoints, guiding the detection network to align cross-modal vessel structures accurately. 

Extensive clinical observations reveal that intrinsic vessel structure similarities exist across different retinal image modalities, but have not been explored for MRIRG from the perspective of keypoint topology constraints. As presented in Fig.~\ref{fig:problem}, we observe that vessel structure similarities between a fundus image and an FA image, such as locations, boundaries, and vessel shape skeleton, provide significant support for the precision of cross-modal vessel structure alignment. 
Conventional two-step registration frameworks implicitly establish keypoint topology constraints based on the segmented vessel structure. In contrast, \textbf{we argue that these shared similarity characteristics can be exploited to construct keypoint topology constraints through a one-step registration framework}, which employs vessel structures that belong to one retinal image modality to guide keypoint topology constraint construction in another retinal image modality.
A plausible reason for this motivation is that topological relationships among vessel structures (e.g., bifurcations and crossovers) significantly affect the quality of keypoint topology constraint construction.

\begin{table*}[!t]
    \centering
    \caption{Summary of existing retinal image registration datasets and MICRO. LSFG: Laser Speckle Flowgraphy, CF: Color Fundus, EMA: Erythrocyte-mediated Angiography, OCT/A: Optical Coherence Tomography/Angiography.}
    \label{tab:datasetreview}

    \renewcommand{\arraystretch}{1}
    \setlength{\tabcolsep}{2.3pt}
    \setlength{\aboverulesep}{1pt}
    \setlength{\belowrulesep}{1.5pt}
    \setlength{\extrarowheight}{0pt}

    \begin{tabular}{cccccccc}
        \toprule
        \multicolumn{2}{c}{\textbf{Dataset}} & \textbf{Image Modality} & \textbf{Pairs (Eyes)} & \textbf{Surgical Stages} & \textbf{Diseases} & \textbf{Availability} & \textbf{Publish time} \\
        \midrule
        \multicolumn{8}{c}{\textbf{Single modality}} \\
        \midrule
        \multicolumn{2}{c}{FIRE~\cite{hernandez2017fire}} & CF & 134 (39) & \xmark & - - & Public & 2017 \\
        \multicolumn{2}{c}{FLoRI21~\cite{ding2021flori21}} & UWF FA & 15 (15) & \xmark & - - & Public & 2021 \\
        \multicolumn{2}{c}{LSFG~\cite{sivaraman2025retinaregnet}} & LSFG & 15 (15) & \xmark & Uveal Melanoma & Private & 2024 \\
        \multicolumn{2}{c}{COph100~\cite{hu2025coph100}} & RetCam & 491 (100) & \xmark & ROP & Public & 2025 \\
        \midrule
        \multicolumn{8}{c}{\textbf{Dual modalities}} \\
        \midrule
        \multicolumn{2}{c}{Lee~\cite{lee2015registration}} & CF-OCT & 52 (52) & \xmark & - - & Private & 2015 \\
        \multicolumn{2}{c}{Almasi~\cite{almasi2020registration}} & FA-SLO & 21 (36) & \xmark & Diabetic Retinopathy & Public & 2020 \\
        \multicolumn{2}{c}{PRIME-FP20~\cite{ding2020weakly}} & UWF FA-FP & 15 (15) & \xmark & - - & Public & 2021 \\
        \multicolumn{2}{c}{FOCTAIR~\cite{martinez2021robust}} & FA-OCTA & 172 (29) & \xmark & RVO & On demand & 2021 \\
        \multicolumn{2}{c}{MEMO~\cite{wang2024memo}} & EMA-OCTA & 30 (30) & \xmark & - - & Public & 2024 \\
        \midrule
        \multicolumn{8}{c}{\textbf{Multiple modalities}} \\
        \midrule
        \multicolumn{2}{c}{IR-OCT-OCTA~\cite{sindel2022multi}} & IR OCT OCTA & 134 (46) & \xmark & - - & Private & 2022 \\
        \midrule
        \multirow{3}{*}{\textbf{\makecell{ MICRO \\ (ours)}}} 
            & pre- & CF, FA, ICGA, FAF & 1108 (301) & \xmark & AMD, RVO, HM, ROP & Public & 2026 \\
            & pre-\&intra- & CF, UWF, surgical frame & 207 (10) & \checkmark & retinal diseases & Public & 2026 \\
            & pre-\&post- & CF, UWF, FA & 50 (50) & \checkmark & cataract, ROP, FEVR & Public & 2026 \\
        \bottomrule
    \end{tabular}
\end{table*}

Motivated by the above analysis, we propose a novel one-step registration framework, named One-step Frequency-Structure Similarity Registration Learning ($\text{OFS}^{2}\text{Reg}$) for MRIRG based on a shared detection network, by fully leveraging the potential of cross-modal vessel structure similarities to construct keypoint topology constraints. To be specific, $\text{OFS}^{2}\text{Reg}$ applies a well-designed frequency-guided structure preservation (FGSP) descriptor that integrates a frequency-aware dynamic kernel generator, a high-frequency enhancer, and a low-frequency aggregator. This combination effectively preserves vessel structures in descriptors of corresponding retinal images, facilitating the construction of coarse keypoint topology constraints. Moreover, $\text{OFS}^{2}\text{Reg}$ develops a novel cross-modal vessel structure similarity learning method ($\text{CMVS}^{2}$). By fully mining the merits of cross-modal vessel structure similarity prior through a single-modal stable keypoint mining strategy, a bidirectional geometric projection mechanism, and cross-modal triplet learning, $\text{CMVS}^{2}$ refines keypoint topology constraints to achieve precise alignment of cross-modal vessel structures. In summary, the main contributions of this paper are as follows:
\begin{itemize} 
\item We construct the MICRO benchmark with 1,365 cross-modal retinal image pairs, covering pre-, intra-, and post-operative phases of ophthalmic surgery. To advance the development of this field, the MICRO dataset, code, and comparable deep registration methods will be available on GitHub\footnote{\url{https://github.com/Martin-GD/MICRO-Benchmark}}.

\item  We rethink the MRIRG task from the perspective of keypoint topology constraints, offering an alternative means for one-step deep registration technique design.

\item We propose a One-step Frequency-Structure Similarity Registration Learning ($\text{OFS}^{2}\text{Reg}$) Framework for MRIRG by fully leveraging cross-modal vessel structure similarities to construct keypoint topology constraints with the aid of well-designed FGSP descriptor and  $\text{CMVS}^{2}$.

\item Extensive experiments on the MICRO dataset demonstrate that our proposed $\text{OFS}^{2}\text{Reg}$ generally achieves state-of-the-art performance across different surgical-phase subsets, with strong OOD generalization on MICRO-pre2post. This paper further investigates the generalization of our method across traditional convolutional neural networks and recent vision foundation models.

\end{itemize}

\section{Related work}

\subsection{Retinal Image Registration Datasets}
As summarized in Table \ref{tab:datasetreview}, existing retinal image registration benchmarks can be broadly categorized into single-modality and multi-modality datasets. Single-modality datasets primarily focus on addressing geometric variations and vessel morphology. The FIRE dataset~\cite{hernandez2017fire} is a representative single-modality registration benchmark comprising 134 retinal fundus image pairs, accounting for three characteristics: the degree of image overlap and the presence/absence of anatomical differences. Other benchmarks have explored pathologies and diverse imaging scales to meet specific retinal image registration requirements. FLoRI21~\cite{ding2021flori21} utilizes ultra-wide-field (UWF) fundus photography to align large-scale anatomical structures by covering $200^{\circ}$ of the retinal surface. LSFG, a private dataset used in ~\cite{sivaraman2025retinaregnet}, introduces complex vessel distortions specific to uveal melanoma by constructing retinal image pairs with tumor-induced deformations. Similarly, COph100~\cite{hu2025coph100} comprises 491 image pairs for retinopathy of prematurity (ROP), highlighting extreme vessel tortuosity and atypical growth patterns in neonatal patients. 
Moreover, existing multi-modality datasets mainly provide complementary vessel structure information from diverse retinal image modalities. 
Specifically, Lee~\cite{lee2015registration} includes CF-OCT pairs characterized by speckle noise and significant intensity variations. 
Almasi~\cite{almasi2020registration} provides FA-SLO pairs that reflect differences in contrast and resolution while considering the curved surface of the retina. To advance wide-field analysis, PRIME-FP20~\cite{ding2020weakly} offers concurrently captured FA and FP image pairs that contain modality-specific noise. FOCTAIR~\cite{martinez2021robust} consists of 172 FA-OCTA pairs specifically focused on the vessel structures of the superficial plexus. MEMO~\cite{wang2024memo} introduces the first collection of EMA-OCTA pairs, which are characterized by a large gap in vessel density between absolute flow data and capillary images. Additionally, IR-OCT-OCTA~\cite{sindel2022multi} utilizes three different retinal image modalities to construct 134 multi-modality image pairs to explore the fusion of infrared and angiographic information.
Here, we summarize that current benchmarks are 
restricted to static and predefined retinal image modality pairs, failing to meet the dynamic nature of MRIRG in ophthalmic surgery. To address this issue, we construct the MICRO dataset, which provides dynamic cross-modal retinal image pairs spanning multiple surgical phases.

\subsection{Deep Registration Methods}
Over the years, deep registration methods have greatly promoted MRIRG and can be broadly classified into two mainstream frameworks: two-step segmentation-then-detection and \textit{one-step} detection. The two-step segmentation-then-detection framework comprises a segmentation step for obtaining well-segmented vessel structures and a detection step for constructing an informative keypoint set. For example, 
Wang ~\textit{et al.}~\cite{wang2021robust} employed a content-adaptive segmentation network to convert multimodal images into consistent vessel representations, followed by a feature network to perform keypoint detection and description on the extracted vessel structures.
Zhang ~\textit{et al.}~\cite{zhang2021two} implemented a hierarchical framework that utilizes segmented vessel masks to detect global transformation parameters, which are subsequently integrated into a deformable registration network for pixel-wise refinement.
VDD-Reg~\cite{wang2024memo} introduced a two-step method that first extracts common vessel structures through semi-supervised segmentation and subsequently performs keypoint detection on the resulting masks to accommodate vessel density discrepancies.
Similarly, Veturi ~\textit{et al.}~\cite{veturi2024eyeliner} presented the EyeLiner pipeline, which masks unstable optic disk regions to isolate peripheral vessel structures, upon which the detection and matching of feature points are performed.
Beyond retinal imaging, vascular structure priors have also been explored in coronary artery analysis. Zhang~\textit{et al.}~\cite{zhang2023anatomy} introduced anatomical and topological constraints to preserve complex coronary vessel structures during segmentation. SPR-Net~\cite{zhang2023spr} further learns structural points by combining vascular geometric features and image contextual information to guide coronary artery registration across systolic and diastolic phases.
Here, we suggest that the segmentation step in two-step segmentation-then-detection framework can be considered as keypoint topology constraint construction. The rationale behind this new perspective is that the segmented vessel structures contain inherent topology relationships among keypoints, which has not been studied in previous works.

\begin{figure*}
    \centering
    \includegraphics[width=1\linewidth]{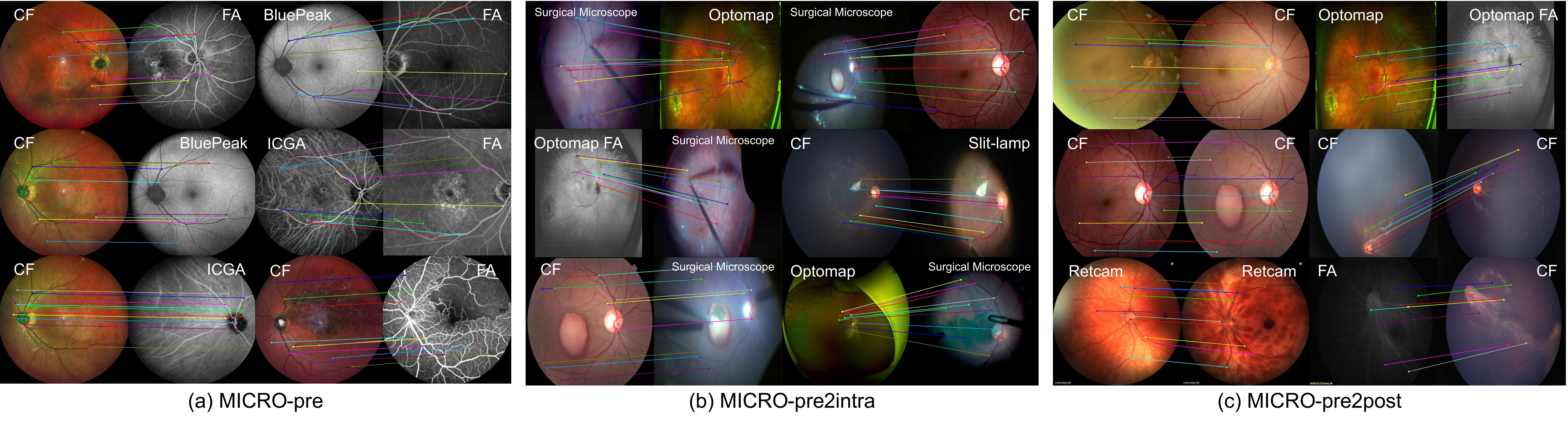}
    \caption{Visualization of representative matching results from the proposed MRIRG dataset, illustrating the diversity of modalities and the complexity of registration tasks. The dataset is categorized into three subsets: (a) \textbf{Pre-operative subset} (MICRO-pre), showing registration between different imaging modalities (e.g., Multicolor, FA, ICGA) with varying fields of view; (b) \textbf{Pre-to-Intra-operative subset} (MICRO-pre2intra), demonstrating the challenging alignment between pre-operative fundus images and intra-operative video frames; and (c) \textbf{Pre-to-Post-operative subset} (MICRO-pre2post), highlighting longitudinal registration across significant temporal gaps. The green lines indicate correct keypoint correspondences established across these challenging cross-modal pairs.}
    \label{fig:dataset_overview}
\end{figure*}

Moreover, the one-step detection framework serves as another prevailing technique of deep registration network design through extracting and matching features from cross-modal retinal image pairs in an end-to-end manner. Truong~\textit{et al.}~\cite{truong2019glampoints} proposed GLAMpoints, a self-supervised framework that performs keypoint detection by identifying stable locations across image transformations. Liu~\textit{et al.}~\cite{liu2022semi} introduced SuperRetina, an end-to-end architecture that jointly performs keypoint detection and description, where a Progressive Keypoint Expansion mechanism is employed to enrich sparse labels along the vessel tree. Similarly, An~\textit{et al.}~\cite{an2022self} implemented a trainable pipeline that extracts vessel-aware convolutional features for joint detection and description. Hu~\textit{et al.}~\cite{hu2023self} exploited the structural consistency of retinal vessels across modalities and introduced structure-preserved modality transformation to support multimodal retinal image registration. Most recently, Wang~\textit{et al.}~\cite{wang2024superjunction} proposed SuperJunction, which incorporates a multi-task branch to perform vessel detection and utilizes constrained negative sampling to localize keypoints at vessel junctions. Although previous one-step detection frameworks have good MRIRG performance by jointly segmenting vessel structures and detecting keypoints, they overlook that similarities between cross-modal vessel structures can be exploited to guide keypoint topology constraint construction without the segmentation step. Therefore, we develop an $\text{OFS}^{2}\text{Reg}$ framework for MRIRG that capitalizes on cross-modal vessel structure similarities for constructing expected keypoint topology constraints. In this paper, we define the keypoint topology constraints in MRIRG under our $\text{OFS}^{2}\text{Reg}$ as follows: leveraging similarities between cross-modal vessel structures to construct keypoint topology for each retinal image modality interactively in a shared detection network.

\section{Dataset}
\label{sec:dataset}

\subsection{Dataset Description}
In this paper, we introduce MICRO, a comprehensive and large benchmark for MRIRG spanning three ophthalmic surgery phases: preoperative planning, intraoperative guidance, and postoperative assessment, as shown in Fig.~\ref{fig:dataset_overview}. The MICRO benchmark comprises 1,365 cross-modal retinal image pairs, curated from both the Beijing Tongren Hospital and publicly available datasets~\cite{timkovivc2024retinal, hu2025coph100}. All clinical data collection was approved by the Ethics Committee of Beijing Tongren Hospital (Approval No.: TREC2022-XJS05), with written informed consent obtained from all participants in adherence to the Declaration of Helsinki. 
Our MICRO consists of three subsets: MICRO-pre, MICRO-pre2intra, and MICRO-pre2post, as describe in the following content.

\subsubsection{MICRO-pre} This subset focuses on MRIRG at the preoperative stage, comprising 1,108 image pairs from 301 eyes (including AMD, RVO, ROP, and HM patients). Multimodal retinal images were collected from diverse medical imaging systems, including CF images, FA images, ICGA, and BluePeak images, as well as RetCam fundus images~\cite{hu2025coph100}. An analysis of this subset, facilitated by control points annotated by experts (detailed in Sec.~\ref{sec:annotation}), reveals a significant average Field of View (FOV) overlap of 64.03\%. Despite the drastic appearance discrepancies shown in Fig.~\ref{fig:dataset_overview}~(a), this high degree of geometric overlap indicates that vessel structures are consistently preserved across modalities. This observation directly motivates our proposed $\text{OFS}^{2}\text{Reg}$ learning framework, as it provides a solid empirical basis for enforcing structural consistency to bridge the modality gap.

\subsubsection{MICRO-pre2intra} This subset contains 207 image pairs linking preoperative planning images (e.g., CF, UWF, slit-lamp images) with intraoperative surgical frames from 10 eyes. These pairs contain four surgical types: vitrectomy, posterior vitreous detachment (PVD) induction, ILM staining, and membrane peeling. Intraoperative frames were extracted from surgical videos with substantial variation between frames, averaging $14.4 \pm 10$ frames per video. 
With only 28.22\% average FOV overlap and interfering surgical instruments, this scenario demands high algorithm robustness for dynamic clinical applications.

\subsubsection{MICRO-pre2post} This subset comprises 50 image pairs from 50 eyes, covering conditions such as cataracts, ROP, and FEVR. Each pair consists of one image captured prior to surgery or laser treatment and a corresponding image acquired during postoperative follow-up. This subset presents a diverse range of postoperative imaging intervals, facilitating the evaluation of treatment efficacy and longitudinal vessel changes. 
This subset has an average FOV overlap of 69.01\% and contains morphological variations associated with surgery or disease progression, providing a challenging setting for evaluating longitudinal registration performance.

\subsection{Manual Annotation Protocol and Benchmark Partitioning}
\label{sec:annotation}
\subsubsection{Manual Annotation Protocol}

Four clinicians participated in the annotation and quality-control process: three clinicians with more than five years of experience in ophthalmic image interpretation, and one senior clinician responsible for quality control. For each cross-modal retinal image pair, 10 corresponding control-point pairs were manually annotated at stable vessel intersections or bifurcations, which are used to estimate the ground-truth rigid transformation $T^{GT}$ between the reference and floating images. During cross-modal training, $T_{A\to B}$ denotes the corresponding $T^{GT}$ from modality $A$ to modality $B$ and provides geometric supervision for BGPM and CMTL.
The ground-truth rigid transformation $T^{GT}$ is estimated from the 10 manually annotated control-point correspondences using the Least-Median-of-Squares (LMEDS) estimator, which reduces the influence of outlier correspondences. The same procedure is applied to all modality pairs and all three MICRO subsets.

\begin{figure}
    \centering
    \includegraphics[width=1\linewidth]{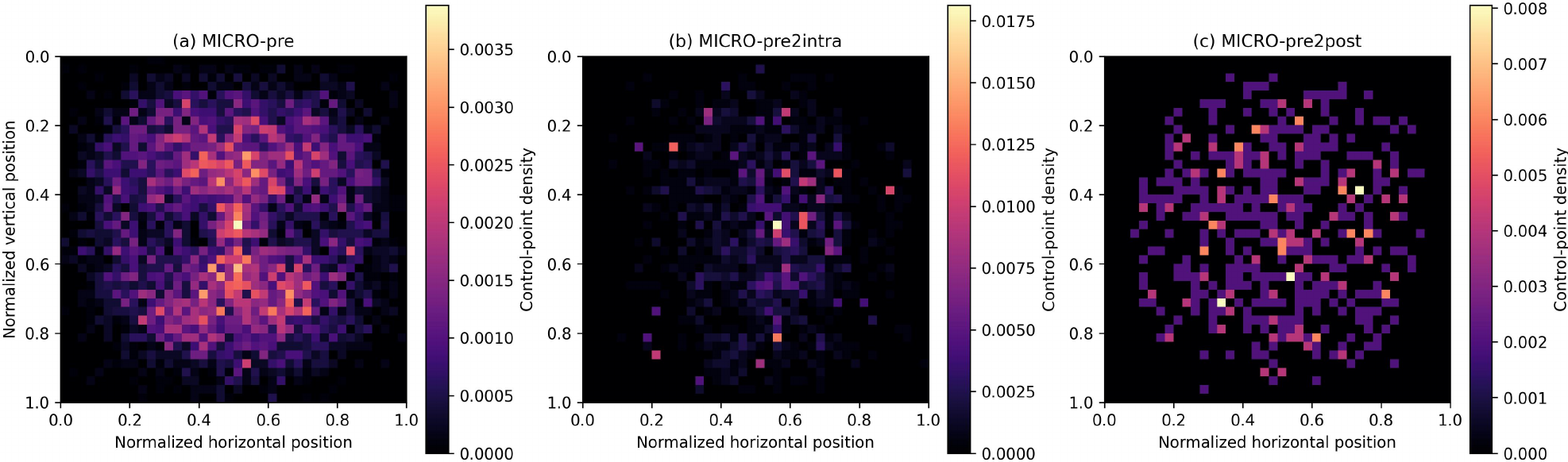}
    \caption{Normalized spatial distributions of manually annotated control points across the retinal field of view in (a) MICRO-pre, (b) MICRO-pre2intra, and (c) MICRO-pre2post.}
    \label{fig:control_point_distribution}
\end{figure}

Let $E_{1j}$ denote the spatial coordinates of the $j$-th control point in the reference image, and $E_{ij}$ denote its position in the floating image. For registration effect evaluation, $E_{ij}$ is mapped to $M_{ij}$ using the estimated transformation, and the registration error is measured by the Euclidean distance between $M_{ij}$ and $E_{1j}$, where a smaller distance indicates better spatial alignment. 
To ensure annotation reliability, we calculate the rigid fitting error after estimating $T^{GT}$ from the 10 manually annotated control-point correspondences as
\begin{equation}
    \mathrm{RMSE}_{GT}
    =
    \sqrt{
    \frac{1}{10}
    \sum_{j=1}^{10}
    \left\|
    T^{GT}(E_{ij})-E_{1j}
    \right\|_2^2
    }.
    \label{eq:gt_rmse}
\end{equation}
Annotations with $\mathrm{RMSE}_{GT}>5$ pixels are reviewed and corrected by the senior clinician until the criterion is satisfied. As shown in Table~\ref{tab:annotation_quality}, the final $\mathrm{RMSE}_{GT}$ values are $2.15\pm1.45$, $2.76\pm1.79$, and $2.44\pm1.65$ pixels for MICRO-pre, MICRO-pre2intra, and MICRO-pre2post, respectively.
We evaluate inter-observer and intra-observer variability on 30 representative image pairs sampled from the three MICRO subsets. For each pair, the same 10 vessel intersections or bifurcations were annotated, and the Euclidean distance between repeated annotations was used to measure localization variability. As summarized in Table~\ref{tab:annotation_variability}, the intra-observer and inter-observer errors were 3.41 $\pm$ 2.56 and 3.68 $\pm$ 2.81 pixels, respectively. These results show good reproducibility and consistency of the manual control-point annotations across repeated annotations and different observers.
We also analyze the normalized spatial distributions of all control points across the retinal field of view. As shown in Fig.~\ref{fig:control_point_distribution}, the annotations cover both central and peripheral retinal regions.

\begin{table}[t]
\centering
\caption{FOV overlap and ground-truth rigid fitting errors of the three MICRO subsets.}
\label{tab:annotation_quality}
\renewcommand{\arraystretch}{1}
\setlength{\tabcolsep}{4pt}
\begin{tabular}{lccc}
\toprule
\textbf{Subset} &
\textbf{Pairs} &
\textbf{FOV Overlap (\%)} &
\textbf{RMSE$_{GT}$ (px)} \\
\midrule
MICRO-pre       & 1108 & 64.03 & 2.15 $\pm$ 1.45 \\
MICRO-pre2intra & 207  & 28.22 & 2.76 $\pm$ 1.79 \\
MICRO-pre2post  & 50   & 69.01 & 2.44 $\pm$ 1.65 \\
\bottomrule
\end{tabular}
\end{table}

\begin{table}[t]
\centering
\caption{Inter-observer and intra-observer variability of manual control-point localization.}
\label{tab:annotation_variability}
\renewcommand{\arraystretch}{1}
\setlength{\tabcolsep}{4pt}
\begin{tabular}{lccc}
\toprule
\textbf{Variability} &
\textbf{Pairs} &
\textbf{Correspondences} &
\textbf{Error (px)} \\
\midrule
Intra-observer & 30 & 300 & 3.41 $\pm$ 2.56 \\
Inter-observer & 30 & 300 & 3.68 $\pm$ 2.81 \\
\bottomrule
\end{tabular}
\end{table}

\subsubsection{Benchmark Partitioning}
This paper splits the MICRO-pre subset into three patient-level subsets in a 6:1:3 ratio for training, validation, and testing, ensuring that all image pairs from the same patient are assigned to the same subset. The three corresponding subsets contain 180, 30, and 91 eyes, corresponding to 647, 134, and 327 image pairs, respectively.
Moreover, to assess the zero-shot generalization or out-of-distribution generalization capability of all registration methods, the MICRO-pre2intra and MICRO-pre2post subsets are strictly reserved as independent, hold-out testing sets. These subsets are entirely excluded from the training phase, serving as a challenging benchmark to evaluate the model's robustness against complex intraoperative obstructions and postoperative pathological changes.

\begin{figure*}[!t]
    \centering
    \includegraphics[width=1\linewidth]{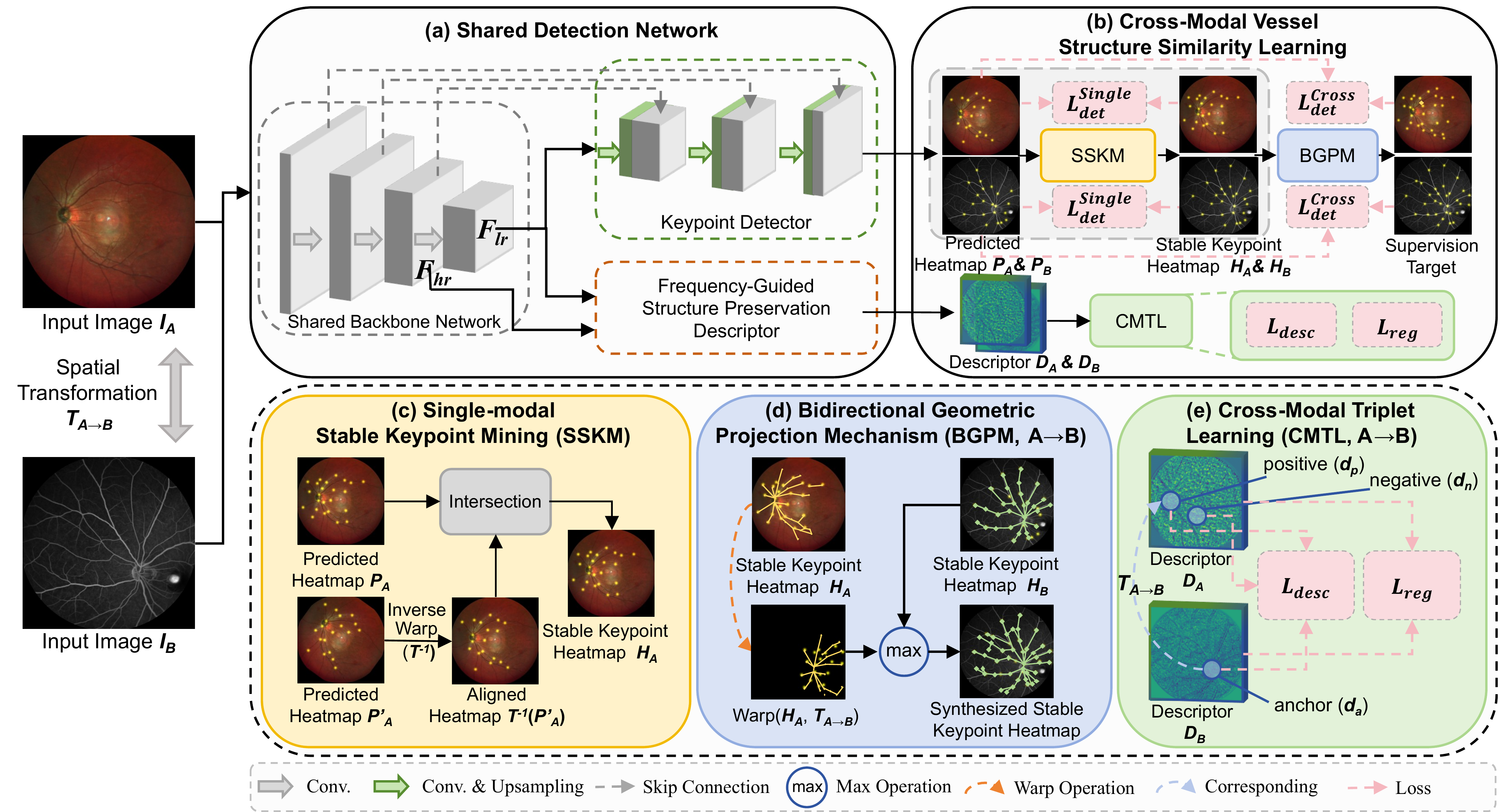}
    \caption{Overview of $\text{OFS}^{2}\text{Reg}$ framework for multimodal retinal image rigid registration, which applies a shared backbone network to produce coarse feature maps, followed by a keypoint detector for generating keypoints, and a frequency-guided structure preservation (FGSP) descriptor for generating distinctive and  modality-invariant descriptors by highlighting significant vessel structures, which can be guided to construct coarse vessel structure topology constraints. Moreover, we employ $\text{CMVS}^{2}$ learning method to refine cross-modal keypoint topology constraints.}
    \label{fig:structure}
\end{figure*}

\section{Methodology}

\subsection{Overview of $\text{OFS}^{2}\text{Reg}$ Framework}
In this paper, we develop the one-step frequency-structure similarity registration learning framework ($\text{OFS}^{2}\text{Reg}$) for MRIRG, as shown in Fig.~\ref{fig:structure}. $\text{OFS}^{2}\text{Reg}$ consists of three core components: a shared detection network, a frequency-guided structure preservation (FGSP) descriptor, and a cross-modal vessel structure similarity ($\text{CMVS}^{2}$) learning method. 
Here, we illustrate the general flowchart of the proposed $\text{OFS}^{2}\text{Reg}$ framework at the training stage as follows:

\begin{itemize}
\item \textbf{Shared detection network}. Given the cross-modal retinal image pairs as the inputs, the shared detection network generates coarse feature maps with varying resolutions for each retinal image modality.

\item \textbf{Frequency-guided structure preservation descriptor}. The FGSP descriptor processes the high- and low-resolution coarse feature maps produced by the shared detection network to emphasize significant vessel structures for establishing vessel structure topology constraints in descriptors with the aid of a frequency-aware dynamic kernel generator, a high-frequency enhancer, and a low-frequency aggregator.

\item \textbf{Cross-modal vessel structure similarity learning method}. $\text{CMVS}^{2}$ first utilizes a single-modal stable keypoint mining strategy to mine more stable keypoints from predicted heatmaps. Subsequently, a bidirectional geometric projection mechanism is applied to refine keypoint topology constraints by fully leveraging the potential of cross-modal vessel structure similarities. Finally, a cross-modal triplet learning strategy is employed to achieve precise cross-modal keypoint matching through optimizing the FGSP descriptor.

\end{itemize}

\begin{figure}
    \centering
    \includegraphics[width=\linewidth]{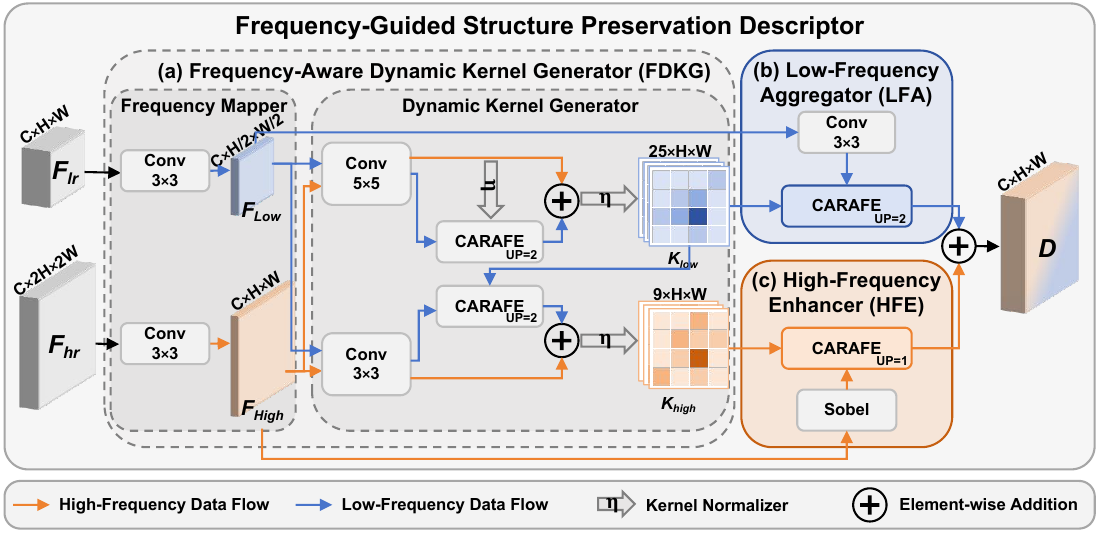}
    \caption{Overview of the Frequency-Guided Structure Preservation (FGSP) descriptor.}
    \label{fig:fgsp}
\end{figure}

\subsection{Frequency-Guided Structure Preservation Descriptor}
\label{subsec:fgsp} 
To construct coarse vessel structure topology constraints in descriptors by emphasizing significant vessel structures, including locations, boundaries, vessel shape skeletons, we develop an FGSP descriptor, which consists of a frequency-aware dynamic kernel generator (FDKG) for generating low- and high-frequency dynamic kernels $K_{low}$ and $K_{high}$, a low-frequency aggregator (LFA) for locating vessel structures and preserving their general shape with the aid of low-frequency dynamic kernels $K_{low}$, and a high-frequency enhancer (HFE) for accentuating ambiguous vessel boundaries and vessel shape skeletons with the aid of high-frequency dynamic kernels $K_{high}$, as shown in Fig.~\ref{fig:fgsp}.

\subsubsection{Frequency-aware Dynamic Kernel Generator} 
As presented in Fig.~\ref{fig:fgsp}(a), given the high- and low-resolution coarse feature maps $F_{hr}\in \mathbb{R}^{C \times 2H \times 2W}$ ($C$ denotes the number of channels, $H$ and $W$ denote the height and width of the feature map) and $F_{lr} \in \mathbb{R}^{C \times H \times W}$ as the inputs, FDKG first applies a frequency mapper to generate low-frequency maps $F_{Low}$ and high-frequency maps $F_{High}$ via two independent convolution operators with kernel size $3 \times 3$:
\begin{equation}
  F_{High} =\text{Conv}_{3\times3}(F_{hr}),  F_{Low} =\text{Conv}_{3\times3}(F_{lr}).
  \label{eq1}
\end{equation}

Next, based on the obtained frequency maps $F_{High} \in \mathbb{R}^{C \times H \times W}$ and $F_{Low} \in \mathbb{R}^{C \times \frac{H}{2} \times \frac{W}{2}}$, we develop a dynamic kernel generator to produce low-frequency dynamic kernels $K_{low} \in \mathbb{R}^{25 \times H \times W}$ and high-frequency dynamic kernels $K_{high} \in \mathbb{R}^{9 \times H \times W}$. The procedures to calculate $K_{low}$ are formulated as follows:
\begin{equation}
\begin{split}
  &F_{H}, F_{L} = \text{Conv}_{5\times5}(F_{High}), \text{Conv}_{5\times5}(F_{Low}), \\
  F_{L}^{up} = &\text{CARAFE}(F_{L}, \eta(F_{H})), ~
  K_{low} = \eta(F_{H} + F_{L}^{up}),
\end{split}
  \label{eq2}
\end{equation}
where $F_{H}, F_{L} \in \mathbb{R}^{25 \times H \times W}$ denote the enhanced high- and low-frequency maps that are obtained via two independent $5 \times 5$ convolutions. $\eta(\cdot)$ denotes the kernel normalizer operator, which is implemented with the combination of a Softmax function and a Hamming window. The key function of kernel normalizer operator is to normalize frequency value distributions, thereby mitigating interference from modality-related noise. 
$\text{CARAFE}(\cdot, \cdot)$~\cite{wang2019carafe} is utilized to upsample $F_L$, where the frequency distribution of the upsampled map $F_L^{up}$ is guided by the enhanced high-frequency information in $F_H$. 
Then, $K_{low}$ denotes the low-frequency dynamic kernels that are obtained with the element-wise addition of $F_{H}$ and $F_{L}^{up}$ and kernel normalizer operator. 

Similarly, we obtain $K_{high}$ with the following equations:
\begin{equation}
    \begin{split}
    &F_{H}', F_{L}' = \text{Conv}_{3\times3}(F_{High}), \text{Conv}_{3\times3}(F_{Low}), \\
    F_{L}^{'up} &= \text{CARAFE}(F_{L}',K_{low} ), ~
    K_{high} = \eta(F_{H}' + F_{L}^{'up}),
    \end{split}
  \label{eq3}
\end{equation}
where $F_{H}', F_{L}' \in \mathbb{R}^{9 \times H \times W}$ are the enhanced high- and low-frequency maps obtained via $3 \times 3$ convolutions. 
$\text{CARAFE}(\cdot, \cdot)$ is utilized to upsample $F_{L}'$, where the upsampled distribution $F_{L}^{'up}$ is guided by the low-frequency dynamic kernels $K_{low}$ obtained from Eq. (\ref{eq2}). 
Then,  $K_{high}$ denotes the high-frequency dynamic kernels that are obtained by applying the $\eta(\cdot)$ operator to the element-wise addition of $F_{H}'$ and $F_{L}^{'up}$.

\subsubsection{Low-Frequency Aggregator} 
As shown in Fig.~\ref{fig:fgsp}(b), the LFA is designed to locate vessel structures and preserve their general shapes, which is essential for ensuring precise coarse alignment of the general retinal structures. Here, we implement LFA with the following equation~\eqref{eq4}:
\begin{equation}
    F_{LFA} = \text{CARAFE}(\mathcal{T}_{low}(F_{Low}), K_{low}),
    \label{eq4}
\end{equation}
where $\mathcal{T}_{low}(\cdot)$ denotes an enhancement operator implemented via a $3 \times 3$ convolution, 
and $\text{CARAFE}(\cdot, \cdot)$ aims to adjust the frequency value distribution guided by the low-frequency dynamic kernels $K_{low}$.

\subsubsection{High-Frequency Enhancer} 
As shown in Fig.~\ref{fig:fgsp}(c), HFE accentuates ambiguous boundaries of vessels and vessel shape skeletons, which is crucial for attaining precise anchors. In this paper, we achieve these goals as follows:
\begin{equation}
    F_{HFE} = \text{CARAFE}(\mathcal{T}_{high}(F_{High}), K_{high}),
    \label{eq5-1}
\end{equation}
where $\mathcal{T}_{high}(\cdot)$ denotes an enhancement operator with help of a Sobel filter to explicitly emphasize structural edges. 
$\text{CARAFE}(\cdot, \cdot)$ aims to refine the feature reassembly guided by the high-frequency dynamic kernels $K_{high}$ to accentuate the delicate geometric details of the vessel bifurcations.

Finally, we obtain the distinctive and modality-invariant descriptors $D$ by element-wise addition of the enhanced high-frequency maps $F_{HFE}$ and the aggregated low-frequency maps $F_{LFA}$. These descriptors include coarse vessel structure topology constraints, which are significant for subsequent cross-modal keypoint matching via cross-modal descriptor loss and cross-modal keypoint topology constraint loss.

\subsection{Cross-modal Vessel Structure Similarity Learning Method} 
\label{subsec:CMVS} 
To effectively refine keypoint topology constraint construction by fully leveraging the merits of cross-modal vessel structure similarities for precise cross-modal keypoint matching, this paper develops the cross-modal vessel structure similarity learning ($\text{CMVS}^{2}$) method, as presented in Fig.~\ref{fig:structure}(b), comprising single-modal stable keypoint mining (SSKM), a bidirectional geometric projection mechanism (BGPM), and a cross-modal triplet learning (CMTL). 
In the following sections, this paper describes our $\text{CMVS}^{2}$ step by step.

\subsubsection{Single-Modal Stable Keypoint Mining}
To effectively select more stable keypoints from the predicted heatmaps for each single retinal image modality, we develop a single-modal stable keypoint mining (SSKM) strategy. As illustrated in Fig.~\ref{fig:structure}(c), SSKM first obtains the predicted heatmaps $P_{A}$ and $P'_{A}$ generated by the keypoint detector, which takes the retinal image $I_{A}$ and its randomly transformed variant $I'_{A} = T(I_A)$ as the inputs, respectively. Next, SSKM conducts the inverse transformation $T^{-1}$ to $P'_{A}$ to generate the spatially aligned heatmap $T^{-1} (P'_{A})$. Finally, a pixel-wise intersection operator is employed to extract coincident high-confidence keypoints from aligned heatmap $T^{-1} (P'_{A})$ and predicted heatmap $P_{A}$ for building the stable keypoint heatmap $\mathcal{H}_{A}$, with the soft-label threshold set to $0.05$ in all experiments. Moreover, to preserve local topological consistency in a single retinal image modality $I_{A}$, the keypoint detector is optimized by a self-supervised detection loss $\mathcal{L}_{det_{A}}$ between the predicted $P_A$ and the mined target $\mathcal{H}_A$ at the single-modal training stage:
\begin{equation}
    \mathcal{L}_{det_{A}}^{Single} = \text{Dice}(P_A, \mathcal{H}_A) + \text{BCE}(P_A, \mathcal{H}_A),
    \label{eq7}
\end{equation}
where $\text{Dice}(P_A, \mathcal{H}_A)$ and $\text{BCE}(P_A, \mathcal{H}_A)$ denote the Dice and Binary Cross-Entropy losses, respectively. Specifically, Dice loss mitigates class imbalance through spatial overlap, while BCE loss ensures pixel-wise probability alignment. Similarly, we perform the above operations to obtain the stable keypoint heatmap $\mathcal{H}_B$ and the corresponding
self-supervised detection loss $\mathcal{L}_{det_{B}}$. We argue that 
this self-supervised constraint learning at the single-modal training stage provides robust single-modal keypoint extraction as a prerequisite for subsequent cross-modal keypoint topology constraint refinement. 

\subsubsection{Bidirectional Geometric Projection Mechanism} 
Following the SSKM strategy, we develop a bidirectional geometric projection mechanism (BGPM) to refine cross-modal keypoint topology constraints by fully leveraging the merits of cross-modal vessel structure similarities. Here, we utilize the stable keypoint heatmap $\mathcal{H}_A$ to refine keypoint topology constraints in the stable keypoint heatmap $\mathcal{H}_B$ as an example to illustrate BGPM in detail, as shown in Fig.~\ref{fig:structure}(d). First, the BGPM maps the stable keypoint heatmap $\mathcal{H}_A$ into the coordinate space of the retinal image modality B via $\text{Warp}(\mathcal{H}_A, \mathcal{T}_{A \to B})$, where $\text{Warp}(\cdot)$ denotes the spatial transformation operator using the ground-truth transformation $\mathcal{T}_{A \to B}$. Here, $\mathcal{T}_{A\to B}$ denotes the ground-truth rigid transformation from image $A$ to image $B$, estimated from the manually annotated control-point correspondences.
Next, the synthesized stable keypoint heatmap $\hat{\mathcal{H}}_B$ is obtained via an element-wise maximum between the stable keypoint heatmap $\mathcal{H}_{B}$ and the projected keypoint topology constraint:
\begin{equation}
    \hat{\mathcal{H}}_B = \max(\underbrace{\mathcal{H}_{B}}_{\text{Intrinsic}}, \underbrace{\text{Warp}(\mathcal{H}_A, \mathcal{T}_{A \to B})}_{\text{Projected Constraint}}).
\end{equation}
The $\max$ operator refines cross-modal keypoint topology constraints by enforcing structural consistency at spatially aligned locations. Moreover, we apply a cross-modal detection loss $\mathcal{L}_{det_{B}}^{Cross}$ to optimize the keypoint detector based on the predicted heatmap $P_B$ and the synthesized stable heatmap $\hat{\mathcal{H}}_B$, to enhance the robustness of keypoint detection against retinal image modality variations and noise:
\begin{equation}
    \mathcal{L}_{det_{B}}^{Cross} = \text{Dice}(P_B, \hat{\mathcal{H}}_B) + \text{BCE}(P_B, \hat{\mathcal{H}}_B).
\end{equation}
Symmetrically, we employ the stable keypoint heatmap $\mathcal{H}_B$ to refine keypoint topology constraints in the stable keypoint heatmap $\mathcal{H}_A$ and calculate the cross-modal detection loss $\mathcal{L}_{det_{A}}^{Cross}$ by following the same procedures in the BGPM.
The objective of the BGPM is to reduce keypoint topological inconsistencies between two retinal image modalities, thereby enabling the model to extract identical topological keypoints.

\subsubsection{Cross-Modal Triplet Learning}

To learn discriminative and modality-invariant descriptors, we develop a cross-modal triplet learning (CMTL) strategy to optimize the FGSP descriptor. As shown in Fig.~\ref{fig:structure}(e), we adopt descriptor $D_A$ to guide descriptor $D_B$ as an example to describe CMTL clearly. We first extract anchor descriptors $\mathbf{d}_a$ from $D_B$ based on the stable keypoint heatmap $\mathcal{H}_B$. Next, their corresponding positive samples $\mathbf{d}_p$ are identified in $D_A$ by mapping the keypoint coordinates via the projection operator $\mathcal{T}_{B \to A}$. Then, we calculate cross-modal guided descriptor loss $\mathcal{L}_{desc_{B}}$ as follows:
\begin{equation}
    \mathcal{L}_{desc_{B}} = \sum_{k \in \{h, r\}} \max(0, \|\mathbf{d}_a - \mathbf{d}_p\| - \|\mathbf{d}_a - \mathbf{d}_{n,k}\| + m),
\end{equation}
where $\mathcal{L}_{desc_{B}}$ comprises two triplet terms involving a random negative $\mathbf{d}_{n,r}$ and a hard negative $\mathbf{d}_{n,h}$, with the triplet margin set to $m=0.8$ in all experiments.
Specifically, $\mathbf{d}_{n,r}$ is randomly sampled from non-corresponding descriptors to provide a contrastive reference. 
In contrast, the hard negative descriptor $\mathbf{d}_{n,h}$ is identified by searching for the descriptor with the shortest distance to the anchor:
\begin{equation}
    \mathbf{d}_{n,h} = \text{arg}\min_{\mathbf{d}_j \in \Omega} \|\mathbf{d}_a - \mathbf{d}_j\|_2,
\end{equation}
where $\Omega$ denotes the set of all descriptors excluding the positive sample. 
Moreover, we apply the cross-modal keypoint topology order to  construct the cross-modal keypoint topology constraint loss $\mathcal{L}_{reg}$ by fully leveraging consistent keypoint coordinates and their corresponding descriptors:
\begin{equation}
    \mathcal{L}_{reg} = \frac{1}{N^2} \| G^A - G^B \|^2_F,
\end{equation}
where $N$ denotes the number of keypoints in stable keypoint heatmap $\mathcal{H}_B$ and the similarity matrices $G^A$ and $G^B$ are constructed using the $N$ pairs of descriptors $\{\mathbf{d}_{a,i}, \mathbf{d}_{p,i}\}_{i=1}^N$ to capture the structural topology in each retinal image modality:
\begin{equation}
    G^A_{ij} = \mathbf{d}_{p,i}^\top \mathbf{d}_{p,j}, \quad G^B_{ij} = \mathbf{d}_{a,i}^\top \mathbf{d}_{a,j}.
\end{equation}

The similarity matrices $G^A$ and $G^B$ capture the keypoint topology in each retinal image modality, while $\mathcal{L}_{reg}$ minimizes the discrepancy between the two similarity matrices. Therefore, a smaller $\mathcal{L}_{reg}$ indicates stronger cross-modal keypoint topology consistency.
Similarly, by applying the CMTL strategy with $D_B$ guiding $D_A$, we can also obtain $ \mathcal{L}_{desc_{A}}$. $\mathcal{L}_{reg}$ is inherently symmetric and independent of the supervision direction.

Furthermore, the cross-modal detection loss, cross-modal guided descriptor loss, and cross-modal keypoint topology constraint loss obtained in BGPM and CMTL constitute the total loss to optimize our proposed framework at the cross-modal training stage, including $\mathcal{L}_{total}^{A \to B}$ and $\mathcal{L}_{total}^{B \to A}$, which are formulated as follows:
$ \mathcal{L}_{total}^{A \to B}= \lambda_{det}\mathcal{L}_{det_{B}}^{Cross} + \lambda_{desc}\mathcal{L}_{desc_{B}} + \lambda_{reg}\mathcal{L}_{reg}$ and 
$ \mathcal{L}_{total}^{B \to A}= \lambda_{det}\mathcal{L}_{det_{A}}^{Cross} + \lambda_{desc}\mathcal{L}_{desc_{A}} + \lambda_{reg}\mathcal{L}_{reg}$. 
Here, $\lambda_{det}$ weights the cross-modal detection loss, which constrains the detector to preserve stable keypoint locations across modalities; $\lambda_{desc}$ weights the cross-modal guided descriptor loss for learning discriminative and modality-invariant descriptors; and $\lambda_{reg}$ weights the cross-modal keypoint topology constraint loss to preserve consistent keypoint topology across modalities. In all experiments, we set $\lambda_{det}=0.5$, $\lambda_{desc}=1.0$, and $\lambda_{reg}=0.1$.

\section{Experiments}
\subsection{Experimental Settings}
\subsubsection{Implementation Details}
We implement $\text{OFS}^{2}\text{Reg}$ and competitive registration methods with PyTorch framework and Python. All experiments are run on an NVIDIA RTX A6000 GPU. We adopt a SuperPoint-like encoder~\cite{detone2018superpoint} as the shared detection network. 
Retinal images are resized to $768\times768$ before both training and evaluation, and all pixel-level registration errors are measured in the resized image space. The maximum number of keypoints is set to 1024. Specific-designed matching modules are retained when available; otherwise, KNN matching is used. The Least-Median-of-Squares (LMEDS) estimator is used to estimate rigid transformations.
The Adam optimizer is employed with a constant learning rate of $10^{-4}$ and a batch size of 2. The training epochs number is set to 100.

\subsubsection{Evaluation Metrics}
Following~\cite{liu2022semi}, this paper assesses the MRIRG performance of our $\text{OFS}^{2}\text{Reg}$ and competitive methods by categorizing registration results into three levels: \textit{failed}, \textit{inaccurate}, and \textit{acceptable}. A case is \textit{failed} if fewer than four matches are detected. For successful matches, we assess accuracy using the Median Error (MedE) and Maximum Error (MaxE) between the annotated control points of the query and reference image. A registration is deemed \textit{acceptable} if it satisfies $\text{MedE} < 20$ and $\text{MaxE} < 50$. Here, we formulate the following metrics:

\begin{itemize}
    \item \textbf{Success Rate (SR)}: The rate of \textit{acceptable} registrations across the total dataset.
    
    \item \textbf{RMSE}: For \textit{acceptable} pairs, the Root Mean Square Error is calculated to measure the pixel-level alignment precision between the warped query points and their reference counterparts.

     \item \textbf{MedE}: For \textit{acceptable} pairs, the Median Error is reported to characterize the typical pixel-level alignment error between the warped query points and their reference counterparts.

    \item \textbf{P95}: For \textit{acceptable} pairs, the 95th Percentile Error is reported to characterize the tail of the registration error distribution and reflect performance on relatively challenging cases.
 
    \item \textbf{mAUC}: Following the evaluation protocol in SuperRetina~\cite{liu2022semi}, we report the mean Area Under the Curve (mAUC) to assess registration robustness. Specifically, mAUC is computed by averaging the success rates over a progressive range of error thresholds from 1 to 25 pixels.
 
\end{itemize}
Statistical significance is conducted using a two-sided Wilcoxon signed-rank test on pair-level mean control-point errors between $\text{OFS}^{2}\text{Reg}$ and each competitive method. Inaccurate registrations with valid errors are included, while pairs without a valid registration error for either method are excluded from the paired comparison.

\subsection{Ablation Study}
In this section, we conduct a series of ablation experiments on the MICRO-pre dataset to investigate the effectiveness of FGSP descriptor and $\text{CMVS}^{2}$ in our $\text{OFS}^{2}\text{Reg}$. SuperRetina~\cite{liu2022semi} serves as our baseline, which adopts a four-stage ResNet-style~\cite{he2016deep} encoder as the shared detection network for joint keypoint detector and descriptor learning.

\begin{table}[!t]
\centering
\caption{Performance comparisons of FGSP Descriptor and $\text{CMVS}^{2}$. 
}
\label{tab:ablation_main}
\renewcommand{\arraystretch}{1}
\setlength{\tabcolsep}{3pt}

\resizebox{0.98\columnwidth}{!}{%
\begin{tabular}{lccccc}
    \toprule
    \multirow{2}{*}{Method} &
    \multicolumn{2}{c}{Component} &
    \multicolumn{3}{c}{Metrics} \\
    & FGSP & $\text{CMVS}^{2}$ &
    RMSE $\downarrow$ & mAUC $\uparrow$ & SR $\uparrow$ \\
    \midrule
    
    Baseline
    & -- & --
    & $6.14 \pm 0.56$
    & $0.352 \pm 0.018$
    & $0.429 \pm 0.014$ \\

    + FGSP
    & \checkmark & --
    & $6.32 \pm 0.29$
    & $0.410 \pm 0.020$
    & $0.541 \pm 0.036$ \\

    + $\text{CMVS}^{2}$
    & -- & \checkmark
    & $5.42 \pm 0.28$
    & $0.596 \pm 0.012$
    & $0.750 \pm 0.004$ \\

    \textbf{$\text{OFS}^{2}\text{Reg}$}
    & \checkmark & \checkmark
    & $\mathbf{4.84 \pm 0.31}$
    & $\mathbf{0.669 \pm 0.025}$
    & $\mathbf{0.806 \pm 0.026}$ \\

    \bottomrule
\end{tabular}%
}
\end{table}

\begin{table}[!t]
\centering
\caption{Keypoint topology consistency comparisons of FGSP Descriptor and $\text{CMVS}^{2}$ on the MICRO-pre subset.
}
\label{tab:ablation_ir}
\renewcommand{\arraystretch}{1}
\setlength{\tabcolsep}{3pt}

\resizebox{0.98\columnwidth}{!}{%
\begin{tabular}{lccccc}
    \toprule
    \multirow{2}{*}{Method} &
    \multicolumn{2}{c}{Component} &
    \multicolumn{3}{c}{Mean Pair Inlier Ratio (MPIR)} \\
    & FGSP & $\text{CMVS}^{2}$ &
    MPIR@5 $\uparrow$ & MPIR@10 $\uparrow$ & MPIR@15 $\uparrow$ \\
    \midrule
    
    Baseline
    & -- & --
    & $0.158 \pm 0.006$
    & $0.404 \pm 0.012$
    & $0.510 \pm 0.015$ \\

    + FGSP
    & \checkmark & --
    & $0.174 \pm 0.006$
    & $0.354 \pm 0.006$
    & $0.448 \pm 0.002$ \\

    + $\text{CMVS}^{2}$
    & -- & \checkmark
    & $0.173 \pm 0.004$
    & $0.415 \pm 0.008$
    & $0.556 \pm 0.006$ \\

    \textbf{$\text{OFS}^{2}\text{Reg}$}
    & \checkmark & \checkmark
    & $\mathbf{0.189 \pm 0.006}$
    & $\mathbf{0.439 \pm 0.002}$
    & $\mathbf{0.577 \pm 0.005}$ \\

    \bottomrule
\end{tabular}%
}
\end{table}

\begin{figure}
\centering
\includegraphics[width=0.98\columnwidth,height=4cm]{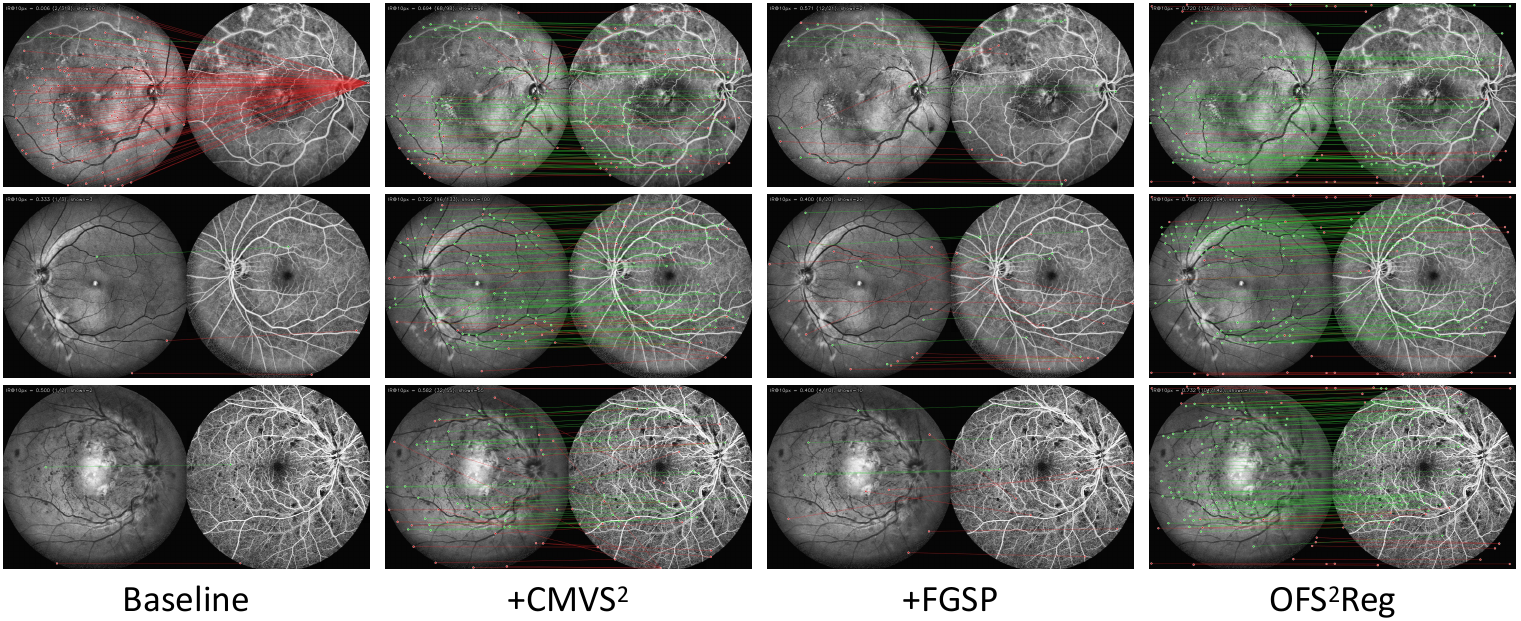}
\caption{Qualitative comparisons of keypoint topology consistency on the MICRO-pre subset. The columns correspond to Baseline, +$\text{CMVS}^{2}$, +FGSP, and $\text{OFS}^{2}\text{Reg}$ from left to right. Green and red lines denote geometrically consistent and inconsistent correspondences, respectively. +$\text{CMVS}^{2}$ reduces inconsistent matches compared with Baseline, while $\text{OFS}^{2}\text{Reg}$ provides the most reliable correspondence patterns.}
\label{fig:keypoint_topology_visual}
\end{figure}

\subsubsection{Effects of FGSP Descriptor and $\text{CMVS}^{2}$ in $\text{OFS}^{2}\text{Reg}$}

Table~\ref{tab:ablation_main} lists the MRIRG results of the baseline, FGSP descriptor, $\text{CMVS}^{2}$, and $\text{OFS}^{2}\text{Reg}$. Compared to baseline, FGSP improves mAUC from 0.352 to 0.410 and SR from 0.429 to 0.541, while the RMSE slightly increases from 6.14 to 6.32 pixels. This is mainly because RMSE is calculated only over acceptable registrations, but the increased SR may introduce more challenging retinal image pairs into the RMSE calculation. This indicates that FGSP mainly improves registration robustness and success rate. It is worth noting that SSKM is retained as the basic training component to ensure the  training ability of the model when evaluating the effects of FGSP. Compared with baseline, $\text{CMVS}^{2}$ reduces RMSE from 6.14 to 5.42 pixels and improves mAUC and SR by 0.244 and 0.321, respectively, showing clear gains in both registration accuracy and robustness. The complete $\text{OFS}^{2}\text{Reg}$  achieves the best RMSE, mAUC, and SR of 4.84, 0.669, and 0.806, respectively, demonstrating the superiority of our method in leveraging cross-modal vessel structure similarities to construct keypoint topology constraints.

To directly evaluate the consistency of cross-modal keypoint topology, we introduce the Mean Pair Inlier Ratio (MPIR). For each predicted correspondence $(p_i,q_i)$, its geometric error is calculated as $e_i=\|T^{GT}(p_i)-q_i\|_2$. A correspondence is regarded as an inlier when $e_i<\tau$, where $\tau$ is set to 5, 10, and 15 pixels. The inlier ratio is first calculated for each image pair and then averaged across all test pairs to obtain MPIR@$\,\tau$.
Table~\ref{tab:ablation_ir} presents quantitative cross-modal keypoint topology consistency comparisons using MPIR. Compared with baseline, $\text{CMVS}^{2}$ improves MPIR@5, MPIR@10, and MPIR@15 from 0.158, 0.404, and 0.510 to 0.173, 0.415, and 0.556, respectively. The complete $\text{OFS}^{2}\text{Reg}$ achieves the highest MPIR@5, MPIR@10, and MPIR@15 of 0.189, 0.439, and 0.577, respectively, demonstrating improved cross-modal keypoint topology consistency. As shown in Fig.~\ref{fig:keypoint_topology_visual}, $\text{CMVS}^{2}$ also improves cross-modal keypoint matching, producing more geometrically consistent correspondences (green lines) through comparisons to baseline. Our $\text{OFS}^{2}\text{Reg}$ further shows a clear qualitative improvement over +FGSP. The visual results demonstrate that $\text{CMVS}^{2}$ effectively improves cross-modal keypoint topology consistency.

\begin{table}[t]
\centering
\caption{Performance comparisons of each component and their combinations in FGSP descriptor.}
\label{tab:ablation_frequency}
\renewcommand{\arraystretch}{1}
\setlength{\tabcolsep}{3.5pt}

\resizebox{0.98\columnwidth}{!}{%
\begin{tabular}{ccc|ccc}
    \toprule
    \multicolumn{3}{c|}{Component} & \multicolumn{3}{c}{Metrics} \\
    LFA & HFE & FDKG & RMSE $\downarrow$ & mAUC $\uparrow$ & SR $\uparrow$ \\
    \midrule

    -- & -- & --
    & $5.42 \pm 0.28$
    & $0.596 \pm 0.012$
    & $0.750 \pm 0.004$ \\

    \checkmark & -- & --
    & $5.20 \pm 0.22$
    & $0.622 \pm 0.017$
    & $0.762 \pm 0.020$ \\

    -- & \checkmark & --
    & $5.24 \pm 0.25$
    & $0.631 \pm 0.024$
    & $0.772 \pm 0.034$ \\

    \checkmark & \checkmark & --
    & $5.09 \pm 0.21$
    & $0.642 \pm 0.019$
    & $0.784 \pm 0.023$ \\
    
    \checkmark & \checkmark & \checkmark
    & $\mathbf{4.84 \pm 0.31}$
    & $\mathbf{0.669 \pm 0.025}$
    & $\mathbf{0.806 \pm 0.026}$ \\

    \bottomrule
\end{tabular}%
}
\end{table}

\subsubsection{Effects of Each Component in FGSP Descriptor}
Table~\ref{tab:ablation_frequency} compares the MRIRG performance of LFA, HFE, FDKG, and their combinations in FGSP descriptor. It can be observed that each component is conducive to performance improvement, and their combinations further enhance MRIRG results. FGSP descriptor with the combination of LFA, HFE, and FDKG gets the best MRIRG performance through comparisons to other combinations.

\subsubsection{Comparison with Multi-Scale Feature Enhancement Methods}
To evaluate the effectiveness of the FGSP descriptor, we construct a multi-scale feature enhancement (MSFE) module based on common multi-scale convolutional designs~\cite{guo2022segnext,li2023large,zhang2025enhanced} for comparisons. MSFE employs parallel convolutional branches with different receptive fields to extract multi-scale spatial features, followed by a $1\times1$ convolution for feature fusion. For a fair comparison, MSFE replaces FGSP at the same position and is trained based on the same experimental settings.

Table~\ref{tab:multiscale_comparison} reports the comparison between FGSP and MSFE under the same $\text{CMVS}^{2}$ setting. It can be observed that MSFE improves the MRIRG performance compared to the single $\text{CMVS}^{2}$. Compared with MSFE, $\text{OFS}^{2}\text{Reg}$ with FGSP further reduces RMSE from 5.07 to 4.84 pixels and improves mAUC and SR by 0.042 and 0.033, respectively. 
These results indicate that FGSP provides more effective feature enhancement than generic multi-scale feature enhancement.

\begin{table}[t]
\centering
\caption{Comparison of FGSP with generic multi-scale feature enhancement on the MICRO-pre subset under the same $\text{CMVS}^{2}$ setting.}
\label{tab:multiscale_comparison}
\renewcommand{\arraystretch}{1}
\setlength{\tabcolsep}{3.5pt}

\resizebox{0.98\columnwidth}{!}{%
\begin{tabular}{lcccc}
\toprule
\textbf{Method} &
\textbf{Params} &
\textbf{RMSE $\downarrow$} &
\textbf{mAUC $\uparrow$} &
\textbf{SR $\uparrow$} \\
\midrule

$\text{CMVS}^{2}$ only
& 3.001M
& $5.42 \pm 0.28$
& $0.596 \pm 0.012$
& $0.750 \pm 0.004$ \\

$\text{CMVS}^{2}$ + MSFE
& 3.049M
& $5.07 \pm 0.25$
& $0.627 \pm 0.022$
& $0.773 \pm 0.030$ \\

$\text{OFS}^{2}\text{Reg}$
& 3.055M
& $\mathbf{4.84 \pm 0.31}$
& $\mathbf{0.669 \pm 0.025}$
& $\mathbf{0.806 \pm 0.026}$ \\

\bottomrule
\end{tabular}%
}
\end{table}

\begin{table}[t]
\centering
\caption{Performance comparisons of each component in $\text{CMVS}^{2}$.}
\label{tab:ablation_CMVS}
\renewcommand{\arraystretch}{1}
\setlength{\tabcolsep}{3.5pt}

\resizebox{0.98\columnwidth}{!}{%
\begin{tabular}{ccc|ccc}
    \toprule
    \multicolumn{3}{c|}{Component} & \multicolumn{3}{c}{Metrics} \\
    SSKM & BGPM & CMTL & RMSE $\downarrow$ & mAUC $\uparrow$ & SR $\uparrow$ \\
    \midrule

    \checkmark & -- & --
    & $6.32 \pm 0.29$
    & $0.410 \pm 0.020$
    & $0.541 \pm 0.036$ \\

    \checkmark & \checkmark & --
    & $5.44 \pm 0.30$
    & $0.616 \pm 0.027$
    & $0.746 \pm 0.020$ \\

    \checkmark & -- & \checkmark
    & $5.06 \pm 0.13$
    & $0.543 \pm 0.033$
    & $0.700 \pm 0.040$ \\

    \checkmark & \checkmark & \checkmark
    & $\mathbf{4.84 \pm 0.31}$
    & $\mathbf{0.669 \pm 0.025}$
    & $\mathbf{0.806 \pm 0.026}$ \\

    \bottomrule
\end{tabular}%
}
\end{table}

\begin{table}[!t]
\centering
\caption{Keypoint topology consistency comparisons of different components in $\text{CMVS}^{2}$ on the MICRO-pre subset.}
\label{tab:ablation_CMVS_ir}
\renewcommand{\arraystretch}{1}
\setlength{\tabcolsep}{3.5pt}

\resizebox{0.98\columnwidth}{!}{%
\begin{tabular}{cccccc}
    \toprule
    \multicolumn{3}{c}{Component} &
    \multicolumn{3}{c}{Mean Pair Inlier Ratio (MPIR)} \\
    SSKM &
    BGPM &
    CMTL &
    MPIR@5 $\uparrow$ &
    MPIR@10 $\uparrow$ &
    MPIR@15 $\uparrow$ \\
    \midrule

    \checkmark & -- & --
    & $0.174 \pm 0.006$
    & $0.354 \pm 0.006$
    & $0.448 \pm 0.002$ \\

    \checkmark & \checkmark & --
    & $0.178 \pm 0.004$
    & $0.374 \pm 0.006$
    & $0.479 \pm 0.004$ \\

    \checkmark & -- & \checkmark
    & $0.165 \pm 0.005$
    & $0.407 \pm 0.002$
    & $0.525 \pm 0.006$ \\
    
    \checkmark & \checkmark & \checkmark
    & $\mathbf{0.189 \pm 0.006}$
    & $\mathbf{0.439 \pm 0.002}$
    & $\mathbf{0.577 \pm 0.005}$ \\

    \bottomrule
\end{tabular}%
}
\end{table}

\subsubsection{Effects of Each Component in  $\text{CMVS}^{2}$}
Table \ref{tab:ablation_CMVS} presents the performance comparisons of various component combinations in $\text{CMVS}^{2}$. 
All $\text{CMVS}^{2}$ ablation experiments are conducted with FGSP enabled. Therefore, the SSKM-only setting in Tables~\ref{tab:ablation_CMVS} and~\ref{tab:ablation_CMVS_ir} corresponds to the FGSP+SSKM configuration and is equivalent to the FGSP setting in Table~\ref{tab:ablation_main}.
It can be observed that BGPM and CMTL are conducive to MRIRG performance gains, and their combinations further enhance MRIRG results. Specifically,  $\text{CMVS}^{2}$ with a combination of SSKM, BGPM, and CMTL achieves the best MRIRG performance, obtaining gains of 0.259 in mAUC and 0.265 in SR compared with using the single SSKM.

Table~\ref{tab:ablation_CMVS_ir} reports the cross-modal keypoint topology consistency results of different components in $\text{CMVS}^{2}$ using MPIR. Compared with SSKM, introducing BGPM consistently improves MPIR across all three thresholds. Introducing CMTL improves MPIR@10 and MPIR@15 to 0.407 and 0.525, respectively. The complete $\text{CMVS}^{2}$ with SSKM, BGPM, and CMTL achieves the best MPIR@5, MPIR@10, and MPIR@15 of 0.189, 0.439, and 0.577, respectively. These results provide direct quantitative evidence that BGPM improves cross-modal keypoint topology consistency and further show the complementary contributions of BGPM and CMTL, with their combination achieving the best overall performance.

\subsubsection{Sensitivity Analysis of Loss Weights}

Table~\ref{tab:hyperparameter_lambda} reports the sensitivity analysis of the three loss weights in $\text{OFS}^{2}\text{Reg}$. We change one loss weight at a time while fixing the other two to their default values. For $\lambda_{det}$, setting it to $0.5$ achieves the best performance across all four metrics, with an MPIR@10 of $0.439$, an RMSE of $4.84$ pixels, an mAUC of $0.669$, and an SR of $0.806$. For $\lambda_{desc}$, the setting of $1.0$ achieves the highest MPIR@10, mAUC, and SR, while $\lambda_{desc}=1.5$ obtains a slightly lower RMSE of $4.81$ pixels. This indicates that $\lambda_{desc}=1.0$ provides a better overall balance between matching accuracy, robustness, and cross-modal keypoint topology consistency. For $\lambda_{reg}$, removing the topology constraint loss by setting $\lambda_{reg}=0$ substantially decreases MPIR@10, mAUC, and SR to $0.382$, $0.573$, and $0.716$, respectively. Increasing $\lambda_{reg}$ to $0.1$ improves all four metrics and achieves the best overall performance, while larger values do not provide further gains. These results provide direct quantitative evidence that $\mathcal{L}_{reg}$ contributes to preserving cross-modal keypoint topology consistency. Based on these results, we set $\lambda_{det}=0.5$, $\lambda_{desc}=1.0$, and $\lambda_{reg}=0.1$ in all experiments.

\begin{table}[!t]
\centering
\caption{Sensitivity analysis of the loss weights in $\text{OFS}^{2}\text{Reg}$ on the MICRO-pre subset.
When varying one loss weight, the other two are fixed to $\lambda_{det}=0.5$, $\lambda_{desc}=1.0$, and $\lambda_{reg}=0.1$.}
\label{tab:hyperparameter_lambda}
\renewcommand{\arraystretch}{1}
\setlength{\tabcolsep}{3.5pt}

\resizebox{0.98\columnwidth}{!}{%
\begin{tabular}{ccccccc}
\toprule
\multicolumn{3}{c}{\textbf{Loss Weight}} &
\multicolumn{4}{c}{\textbf{Metrics}} \\
\cmidrule(lr){1-3}
\cmidrule(lr){4-7}
$\lambda_{det}$ &
$\lambda_{desc}$ &
$\lambda_{reg}$ &
MPIR@10 $\uparrow$ &
RMSE $\downarrow$ &
mAUC $\uparrow$ &
SR $\uparrow$ \\
\midrule

\multicolumn{7}{c}{\textit{\textbf{Sensitivity to $\lambda_{det}$}}} \\
0.1  & 1.0 & 0.1 &
$0.417 \pm 0.005$ &
$5.26 \pm 0.39$ &
$0.633 \pm 0.023$ &
$0.780 \pm 0.029$ \\

0.25 & 1.0 & 0.1 &
$0.423 \pm 0.002$ &
$5.08 \pm 0.24$ &
$0.645 \pm 0.020$ &
$0.798 \pm 0.022$ \\

0.5 & 1.0 & 0.1 &
$\mathbf{0.439 \pm 0.002}$ &
$\mathbf{4.84 \pm 0.31}$ &
$\mathbf{0.669 \pm 0.025}$ &
$\mathbf{0.806 \pm 0.026}$ \\

0.75 & 1.0 & 0.1 &
$0.435 \pm 0.005$ &
$4.93 \pm 0.33$ &
$0.649 \pm 0.033$ &
$0.795 \pm 0.030$ \\

1.0 & 1.0 & 0.1 &
$0.426 \pm 0.004$ &
$5.05 \pm 0.40$ &
$0.652 \pm 0.038$ &
$0.803 \pm 0.045$ \\

\midrule
\multicolumn{7}{c}{\textit{\textbf{Sensitivity to $\lambda_{desc}$}}} \\
0.5 & 0.25 & 0.1 &
$0.411 \pm 0.011$ &
$5.17 \pm 0.44$ &
$0.629 \pm 0.052$ &
$0.780 \pm 0.021$ \\

0.5 & 0.5 & 0.1 &
$0.435 \pm 0.003$ &
$4.88 \pm 0.03$ &
$0.657 \pm 0.022$ &
$0.799 \pm 0.021$ \\

0.5 & 1.0 & 0.1 &
$\mathbf{0.439 \pm 0.002}$ &
$4.84 \pm 0.31$ &
$\mathbf{0.669 \pm 0.025}$ &
$\mathbf{0.806 \pm 0.026}$ \\

0.5 & 1.5 & 0.1 &
$0.432 \pm 0.007$ &
$\mathbf{4.81 \pm 0.35}$ &
$0.652 \pm 0.036$ &
$0.795 \pm 0.037$ \\

0.5 & 2.0 & 0.1 &
$0.425 \pm 0.012$ &
$4.87 \pm 0.25$ &
$0.641 \pm 0.019$ &
$0.784 \pm 0.022$ \\

\midrule
\multicolumn{7}{c}{\textit{\textbf{Sensitivity to $\lambda_{reg}$}}} \\
0.5 & 1.0 & 0 &
$0.382 \pm 0.020$ &
$5.25 \pm 0.35$ &
$0.573 \pm 0.043$ &
$0.716 \pm 0.046$ \\

0.5 & 1.0 & 0.05 &
$0.437 \pm 0.002$ &
$5.00 \pm 0.26$ &
$0.651 \pm 0.017$ &
$0.797 \pm 0.021$ \\

0.5 & 1.0 & 0.1 &
$\mathbf{0.439 \pm 0.002}$ &
$\mathbf{4.84 \pm 0.31}$ &
$\mathbf{0.669 \pm 0.025}$ &
$\mathbf{0.806 \pm 0.026}$ \\

0.5 & 1.0 & 0.2 &
$0.439 \pm 0.006$ &
$4.90 \pm 0.02$ &
$0.664 \pm 0.028$ &
$0.796 \pm 0.020$ \\

0.5 & 1.0 & 0.5 &
$0.431 \pm 0.005$ &
$4.90 \pm 0.36$ &
$0.650 \pm 0.030$ &
$0.786 \pm 0.029$ \\

\bottomrule
\end{tabular}%
}
\end{table}

\begin{table*}[!t]
\centering
\caption{MRIRG performance comparisons of $\text{OFS}^{2}\text{Reg}$ and state-of-the-art methods on the MICRO-pre subset. $^\dagger$P-values are obtained using a two-sided Wilcoxon signed-rank test on pair-level mean control-point errors.}
\label{tab:sota_pre}
\renewcommand{\arraystretch}{1}
\setlength{\tabcolsep}{4pt}
\resizebox{0.95\textwidth}{!}{%
\begin{tabular}{l l ccccccc}
    \toprule
    \textbf{Category} &
    \textbf{Method} &
    \textbf{RMSE $\downarrow$} &
    \textbf{RMSE 95\% CI} &
    \textbf{MedE $\downarrow$} &
    \textbf{P95 $\downarrow$} &
    \textbf{mAUC $\uparrow$} &
    \textbf{SR $\uparrow$} &
    \textbf{P-Value$^\dagger$} \\
    \midrule

    \multirow{2}{*}{\makecell[l]{Traditional \\ registration}}
    & SIFT~\cite{lowe2004distinctive} (2004)
    & 5.32
    & [5.04, 5.60]
    & 3.92
    & 15.67
    & 0.114
    & 0.135
    & $<0.01$ \\

    & GDB-ICP~\cite{yang2007registration} (2007)
    & 6.10
    & [5.71, 6.42]
    & 3.87
    & \underline{13.71}
    & 0.412
    & 0.524
    & $<0.01$ \\

    \midrule

    \multirow{15}{*}{\makecell[l]{Deep \\ registration}}
    & \multicolumn{8}{c}{\textit{\textbf{Two-step:}}} \\

    & EyeLiner~\cite{veturi2024eyeliner} (2024)
    & 6.22
    & [5.96, 6.53]
    & 4.16
    & 14.82
    & 0.501
    & 0.634
    & $<0.01$ \\

    & VDD-Reg~\cite{wang2024memo} (2024)
    & $8.32 \pm 0.35$
    & [7.65, 8.46]
    & $6.39 \pm 0.16$
    & $25.89 \pm 0.68$
    & $0.315 \pm 0.015$
    & $0.459 \pm 0.008$
    & $<0.01$ \\

    \cmidrule(lr){2-9}

    & \multicolumn{8}{c}{\textit{\textbf{One-step:}}} \\

    & SuperPoint~\cite{detone2018superpoint} (2018)
    & $6.44 \pm 0.11$
    & [6.11, 6.67]
    & $4.05 \pm 0.05$
    & $19.47 \pm 0.86$
    & $0.514 \pm 0.009$
    & $0.624 \pm 0.020$
    & $<0.01$ \\

    & GLAMpoints~\cite{truong2019glampoints} (2019)
    & $7.71 \pm 0.13$
    & [7.51, 7.85]
    & $4.83 \pm 0.12$
    & $23.70 \pm 0.62$
    & $0.343 \pm 0.007$
    & $0.469 \pm 0.018$
    & $<0.01$ \\

    & R2D2~\cite{revaud2019r2d2} (2019)
    & $6.65 \pm 0.45$
    & [6.45, 7.01]
    & $3.74 \pm 0.28$
    & $21.62 \pm 3.07$
    & $0.270 \pm 0.039$
    & $0.343 \pm 0.050$
    & $<0.01$ \\

    & SuperRetina~\cite{liu2022semi} (2022)
    & $6.37 \pm 0.26$
    & [5.85, 6.68]
    & $4.36 \pm 0.23$
    & $15.67 \pm 0.49$
    & $0.372 \pm 0.034$
    & $0.505 \pm 0.020$
    & $<0.01$ \\

    & SwinU-SuperRetina~\cite{nasser2024reverse} (2024)
    & $6.42 \pm 0.17$
    & [6.00, 6.84]
    & $4.62 \pm 0.10$
    & $18.84 \pm 0.84$
    & $0.449 \pm 0.014$
    & $0.588 \pm 0.017$
    & $<0.01$ \\

    & LK-SuperRetina~\cite{nasser2024reverse} (2024)
    & $6.49 \pm 0.21$
    & [6.18, 6.65]
    & $4.26 \pm 0.10$
    & $20.18 \pm 0.56$
    & $0.274 \pm 0.019$
    & $0.377 \pm 0.014$
    & $<0.01$ \\

    & SuperJunction~\cite{wang2024superjunction} (2024)
    & \underline{$5.27 \pm 0.08$}
    & [4.97, 5.39]
    & $4.17 \pm 0.15$
    & $13.74 \pm 1.07$
    & $0.382 \pm 0.005$
    & $0.421 \pm 0.009$
    & $<0.01$ \\

    & XFeat~\cite{potje2024xfeat} (2024)
    & $6.43 \pm 0.14$
    & [6.26, 6.68]
    & $5.06 \pm 0.09$
    & $18.60 \pm 0.34$
    & $0.587 \pm 0.011$
    & \underline{$0.771 \pm 0.020$}
    & $<0.01$ \\

    & XFeat*~\cite{potje2024xfeat} (2024)
    & $5.84 \pm 0.13$
    & [5.44, 6.28]
    & $4.11 \pm 0.04$
    & $16.81 \pm 0.44$
    & \underline{$0.602 \pm 0.011$}
    & $0.710 \pm 0.038$
    & $<0.01$ \\

    & CGTME~\cite{yan2025two} (2025)
    & $6.07 \pm 0.07$
    & [5.85, 6.19]
    & \underline{$3.61 \pm 0.07$}
    & $18.44 \pm 0.17$
    & $0.509 \pm 0.028$
    & $0.629 \pm 0.031$
    & $<0.01$ \\

    & \textbf{$\text{OFS}^{2}\text{Reg}$} (Ours)
    & $\mathbf{4.84 \pm 0.31}$
    & [4.46, 5.03]
    & $\mathbf{3.42 \pm 0.002}$
    & $\mathbf{12.84 \pm 0.55}$
    & $\mathbf{0.669 \pm 0.025}$
    & $\mathbf{0.806 \pm 0.026}$
    & -- \\

    \bottomrule
\end{tabular}%
}
\end{table*}

\begin{table}[t]
\centering
\caption{Comparison between the official models and the models retrained on the MICRO-pre training set. Retrained results are reported as mean $\pm$ standard deviation over three independent runs.}
\label{tab:retrain_comparison}
\renewcommand{\arraystretch}{1}
\setlength{\tabcolsep}{3pt}

\resizebox{0.98\columnwidth}{!}{%
\begin{tabular}{llccc}
\toprule

\multirow{2}{*}{\textbf{Method}} &
\multirow{2}{*}{\textbf{Setting}} &
\multicolumn{3}{c}{\textbf{Metrics}} \\
\cmidrule(lr){3-5}
& &
\textbf{RMSE $\downarrow$} &
\textbf{mAUC $\uparrow$} &
\textbf{SR $\uparrow$} \\
\midrule

\multirow{2}{*}{VDD-Reg~\cite{wang2024memo}}
& Official
& 6.90
& 0.334
& 0.428 \\
& MICRO-pre
& $8.32 \pm 0.35$
& $0.315 \pm 0.015$
& $0.459 \pm 0.008$ \\
\midrule

\multirow{2}{*}{SuperPoint~\cite{detone2018superpoint}}
& Official
& 6.48
& 0.507
& 0.634 \\
& MICRO-pre
& $6.44 \pm 0.11$
& $0.514 \pm 0.009$
& $0.624 \pm 0.020$ \\
\midrule

\multirow{2}{*}{GLAMpoints~\cite{truong2019glampoints}}
& Official
& 11.37
& 0.006
& 0.013 \\
& MICRO-pre
& $7.71 \pm 0.13$
& $0.343 \pm 0.007$
& $0.469 \pm 0.018$ \\
\midrule

\multirow{2}{*}{R2D2~\cite{revaud2019r2d2}}
& Official
& 6.67
& 0.384
& 0.496 \\
& MICRO-pre
& $6.65 \pm 0.45$
& $0.270 \pm 0.039$
& $0.343 \pm 0.050$ \\
\midrule

\multirow{2}{*}{SuperRetina~\cite{liu2022semi}}
& Official
& 5.82
& 0.314
& 0.392 \\
& MICRO-pre
& $6.37 \pm 0.26$
& $0.372 \pm 0.034$
& $0.505 \pm 0.020$ \\
\midrule

\multirow{2}{*}{SwinU-SuperRetina~\cite{nasser2024reverse}}
& Official
& 6.56
& 0.401
& 0.520 \\
& MICRO-pre
& $6.42 \pm 0.17$
& $0.449 \pm 0.014$
& $0.588 \pm 0.017$ \\
\midrule

\multirow{2}{*}{LK-SuperRetina~\cite{nasser2024reverse}}
& Official
& 5.94
& 0.261
& 0.330 \\
& MICRO-pre
& $6.49 \pm 0.21$
& $0.274 \pm 0.019$
& $0.377 \pm 0.014$ \\
\midrule

\multirow{2}{*}{SuperJunction~\cite{wang2024superjunction}}
& Official
& 5.12
& 0.354
& 0.431 \\
& MICRO-pre
& $5.27 \pm 0.08$
& $0.382 \pm 0.005$
& $0.421 \pm 0.009$ \\
\midrule

\multirow{2}{*}{XFeat~\cite{potje2024xfeat}}
& Official
& 10.25
& 0.147
& 0.241 \\
& MICRO-pre
& $6.43 \pm 0.14$
& $0.587 \pm 0.011$
& $0.771 \pm 0.020$ \\
\midrule

\multirow{2}{*}{XFeat*~\cite{potje2024xfeat}}
& Official
& 9.39
& 0.165
& 0.250 \\
& MICRO-pre
& $5.84 \pm 0.13$
& $0.602 \pm 0.011$
& $0.710 \pm 0.038$ \\
\midrule

\multirow{2}{*}{CGTME~\cite{yan2025two}}
& Official
& 6.00
& 0.486
& 0.599 \\
& MICRO-pre
& $6.07 \pm 0.07$
& $0.509 \pm 0.028$
& $0.629 \pm 0.031$ \\

\midrule

\textbf{$\text{OFS}^{2}\text{Reg}$} (Ours)
& MICRO-pre
& $\mathbf{4.84 \pm 0.31}$
& $\mathbf{0.669 \pm 0.025}$
& $\mathbf{0.806 \pm 0.026}$ \\

\bottomrule
\end{tabular}%
}
\end{table}

\begin{table*}[!t]
\centering
\caption{OOD MRIRG performance comparisons of $\text{OFS}^{2}\text{Reg}$ and state-of-the-art methods on the MICRO-pre2intra and MICRO-pre2post subsets.}
\label{tab:sota_ood}
\renewcommand{\arraystretch}{1}
\setlength{\tabcolsep}{3.0pt}
\resizebox{\textwidth}{!}{%
\begin{tabular}{l l ccccc @{\hspace{10pt}} ccccc}
    \toprule
    \multirow{2}{*}{\textbf{Category}} &
    \multirow{2}{*}{\textbf{Method}} &
    \multicolumn{5}{c}{\textbf{MICRO-pre2intra}} &
    \multicolumn{5}{c}{\textbf{MICRO-pre2post}} \\
    
    \cmidrule(lr){3-7}
    \cmidrule(l){8-12}
    
    & &
    RMSE $\downarrow$ &
    MedE $\downarrow$ &
    P95 $\downarrow$ &
    mAUC $\uparrow$ &
    SR $\uparrow$ &
    
    RMSE $\downarrow$ &
    MedE $\downarrow$ &
    P95 $\downarrow$ &
    mAUC $\uparrow$ &
    SR $\uparrow$ \\
    \midrule
    
    \multirow{2}{*}{\makecell[l]{Traditional \\ registration}}
    & SIFT~\cite{lowe2004distinctive} (2004)
    & -- & -- & -- & 0.000 & 0.000
    & 8.46 & 6.75 & 26.61 & 0.070 & 0.100 \\
    
    & GDB-ICP~\cite{yang2007registration} (2007)
    & 8.81
    & 5.99
    & 25.67
    & \textbf{0.192}
    & \underline{0.280}
    & 5.08
    & 3.25
    & 16.53
    & 0.560
    & 0.680 \\
    
    \midrule
    
    \multirow{15}{*}{\makecell[l]{Deep \\ registration}}
    & \multicolumn{10}{c}{\textit{\textbf{Two-step:}}} \\
    
    & EyeLiner~\cite{veturi2024eyeliner} (2024)
    & 10.59
    & 7.79
    & 32.00
    & 0.169
    & \textbf{0.333}
    & 5.72
    & 3.03
    & 19.18
    & 0.462
    & 0.560 \\
    
    & VDD-Reg~\cite{wang2024memo} (2024)
    & $8.29 \pm 0.19$
    & $5.80 \pm 0.40$
    & $24.58 \pm 1.23$
    & $0.061 \pm 0.011$
    & $0.082 \pm 0.013$
    & $7.58 \pm 0.21$
    & $4.48 \pm 0.19$
    & $18.05 \pm 0.53$
    & $0.421 \pm 0.019$
    & $0.553 \pm 0.012$ \\
    
    \cmidrule(lr){2-12}
    
    & \multicolumn{10}{c}{\textit{\textbf{One-step:}}} \\
    
    & SuperPoint~\cite{detone2018superpoint} (2018)
    & $9.29 \pm 0.45$
    & $5.86 \pm 0.20$
    & $28.39 \pm 2.40$
    & $0.131 \pm 0.016$
    & $0.182 \pm 0.024$
    & $5.07 \pm 0.59$
    & $3.04 \pm 0.11$
    & $\underline{13.30 \pm 0.54}$
    & $0.568 \pm 0.006$
    & $0.680 \pm 0.020$ \\
    
    & GLAMpoints~\cite{truong2019glampoints} (2019)
    & $11.65 \pm 1.47$
    & $8.01 \pm 1.64$
    & $33.01 \pm 1.52$
    & $0.018 \pm 0.006$
    & $0.032 \pm 0.014$
    & $5.01 \pm 0.62$
    & $\underline{2.96 \pm 0.06}$
    & $14.53 \pm 2.99$
    & $0.527 \pm 0.015$
    & $0.620 \pm 0.020$ \\
    
    & R2D2~\cite{revaud2019r2d2} (2019)
    & $9.06 \pm 0.17$
    & $6.37 \pm 0.33$
    & $24.59 \pm 0.94$
    & $0.165 \pm 0.004$
    & $0.218 \pm 0.018$
    & $4.84 \pm 0.67$
    & $3.10 \pm 0.26$
    & $14.72 \pm 1.91$
    & $0.462 \pm 0.047$
    & $0.540 \pm 0.053$ \\
    
    & SuperRetina~\cite{liu2022semi} (2022)
    & $8.23 \pm 0.76$
    & $5.51 \pm 0.16$
    & $\underline{22.43 \pm 1.24}$
    & $0.073 \pm 0.008$
    & $0.103 \pm 0.012$
    & $5.12 \pm 0.63$
    & $3.11 \pm 0.10$
    & $16.79 \pm 0.67$
    & $0.499 \pm 0.008$
    & $0.573 \pm 0.012$ \\
    
    & SwinU-SuperRetina~\cite{nasser2024reverse} (2024)
    & $\underline{7.93 \pm 0.71}$
    & $5.85 \pm 0.22$
    & $24.75 \pm 3.57$
    & $0.100 \pm 0.004$
    & $0.140 \pm 0.000$
    & $\underline{4.80 \pm 0.19}$
    & $3.41 \pm 0.29$
    & $13.58 \pm 1.45$
    & $0.542 \pm 0.042$
    & $0.633 \pm 0.058$ \\
    
    & LK-SuperRetina~\cite{nasser2024reverse} (2024)
    & $9.10 \pm 2.01$
    & $6.68 \pm 0.10$
    & $38.97 \pm 1.08$
    & $0.032 \pm 0.011$
    & $0.047 \pm 0.012$
    & $4.96 \pm 0.71$
    & $\mathbf{2.93 \pm 0.14}$
    & $19.15 \pm 0.46$
    & $0.515 \pm 0.012$
    & $0.613 \pm 0.012$ \\
    
    & SuperJunction~\cite{wang2024superjunction} (2024)
    & $9.34 \pm 0.51$
    & $6.06 \pm 0.54$
    & $28.57 \pm 2.34$
    & $0.056 \pm 0.007$
    & $0.081 \pm 0.003$
    & $4.84 \pm 0.25$
    & $3.15 \pm 0.10$
    & $13.86 \pm 1.40$
    & $0.503 \pm 0.009$
    & $0.607 \pm 0.042$ \\
    
    & XFeat~\cite{potje2024xfeat} (2024)
    & $10.25 \pm 0.30$
    & $8.25 \pm 0.42$
    & $31.30 \pm 1.71$
    & $0.132 \pm 0.008$
    & $0.216 \pm 0.015$
    & $6.94 \pm 0.68$
    & $4.69 \pm 0.31$
    & $20.15 \pm 2.50$
    & $0.491 \pm 0.033$
    & $0.660 \pm 0.020$ \\
    
    & XFeat*~\cite{potje2024xfeat} (2024)
    & $8.55 \pm 0.64$
    & $6.54 \pm 0.31$
    & $27.28 \pm 0.93$
    & $0.151 \pm 0.015$
    & $0.224 \pm 0.022$
    & $6.11 \pm 0.39$
    & $3.94 \pm 0.16$
    & $16.53 \pm 1.90$
    & $0.434 \pm 0.026$
    & $0.560 \pm 0.053$ \\
    
    & CGTME~\cite{yan2025two} (2025)
    & $9.25 \pm 0.37$
    & $\underline{5.41 \pm 0.22}$
    & $33.13 \pm 1.09$
    & $0.169 \pm 0.003$
    & $0.221 \pm 0.007$
    & $5.31 \pm 0.37$
    & $3.02 \pm 0.07$
    & $15.45 \pm 0.28$
    & $\underline{0.572 \pm 0.006}$
    & $\underline{0.687 \pm 0.023}$ \\
    
    & \textbf{$\text{OFS}^{2}\text{Reg}$} (Ours)
    & $\mathbf{5.23 \pm 0.13}$
    & $\mathbf{5.10 \pm 0.04}$
    & $\mathbf{14.85 \pm 1.60}$
    & $\underline{0.177 \pm 0.001}$
    & $0.229 \pm 0.006$
    & $\mathbf{4.10 \pm 0.16}$
    & $3.10 \pm 0.05$
    & $\mathbf{11.60 \pm 1.20}$
    & $\mathbf{0.632 \pm 0.010}$
    & $\mathbf{0.747 \pm 0.012}$ \\
    
    \bottomrule
\end{tabular}%
}
\end{table*}

\subsection{Comparisons with State-of-the-Art Methods}

To prove the superiority of our $\text{OFS}^{2}\text{Reg}$, we adopt the following SOTA deep registration methods for comparisons: 
\begin{itemize}
    \item Traditional registration: SIFT~\cite{lowe2004distinctive}, GDB-ICP~\cite{yang2007registration}. 
    \item Two-step deep registration: VDD-Reg~\cite{wang2024memo}, EyeLiner~\cite{veturi2024eyeliner}.
    \item One-step deep registration: such as SuperRetina~\cite{liu2022semi}, SuperJunction~\cite{wang2024superjunction}, and CGTME~\cite{yan2025two}.
\end{itemize}

All methods are evaluated on the MICRO benchmark using the same experimental settings. SIFT and GDB-ICP require no training, while EyeLiner follows a training-free paradigm. The deep registration methods are retrained on the MICRO-pre training set. Specific-designed matching modules are retained when available; otherwise, KNN matching is used consistently with $\text{OFS}^{2}\text{Reg}$. All methods trained on the MICRO-pre training set, including $\text{OFS}^{2}\text{Reg}$ and the retrained competitive methods, are evaluated over three independent runs with different random seeds, and the results are reported as mean $\pm$ standard deviation. For RMSE, 95\% confidence intervals (CIs) are estimated using 1,000 bootstrap resamples of the test pairs. For these methods, the same resampled pair indices are used across the three runs, and the mean RMSE across runs is calculated for each resample, while methods evaluated in a single run are bootstrapped directly. Here, RMSE is calculated only over acceptable registrations in each resample.

\subsubsection{Result Comparisons on MICRO-pre}

Table~\ref{tab:sota_pre} reports the MRIRG performance on the MICRO-pre subset. It can be observed that $\text{OFS}^{2}\text{Reg}$ achieves the best performance across all five evaluation metrics. Compared with the traditional GDB-ICP~\cite{yang2007registration}, $\text{OFS}^{2}\text{Reg}$ reduces RMSE from $6.10$ to $4.84$ pixels and improves mAUC by $0.257$. $\text{OFS}^{2}\text{Reg}$ improves mAUC and SR by $0.168$ and $0.172$ through comparisons to the two-step EyeLiner~\cite{veturi2024eyeliner}. Furthermore, among all one-step deep registration methods, $\text{OFS}^{2}\text{Reg}$ reduces RMSE from $5.27$ to $4.84$ pixels compared with SuperJunction~\cite{wang2024superjunction}, while achieving lower MedE and P95 of $3.42$ and $12.84$ pixels, respectively. Compared with XFeat*~\cite{potje2024xfeat} and XFeat, $\text{OFS}^{2}\text{Reg}$ improves mAUC by $0.067$ and SR by $0.035$. Compared with the recent CGTME~\cite{yan2025two}, our $\text{OFS}^{2}\text{Reg}$ improves mAUC and SR by $0.160$ and $0.177$, respectively. In addition, the Wilcoxon signed-rank tests on pair-level mean control-point errors yield $p<0.01$ for all comparisons between $\text{OFS}^{2}\text{Reg}$ and the competitive methods, showing that there are statistical significances between our $\text{OFS}^{2}\text{Reg}$ and other competitive methods.

Table~\ref{tab:retrain_comparison} presents the MRIRG performance of deep registration methods based on the official pre-trained weights and the retrained deep registration methods on the MICRO-pre training set. It sees that most retrained deep registration methods improve MRIRG performance at least one evaluation metric. In particular, GLAMpoints, XFeat, and XFeat* achieve significant MRIRG performance gains after retraining, with RMSE reductions of $3.66$, $3.82$, and $3.55$ pixels, mAUC improvements of $0.337$, $0.440$, and $0.437$, and SR improvements of $0.456$, $0.530$, and $0.460$, respectively. These results demonstrate that retraining deep registration methods on the MICRO-pre training set can effectively improve MRIRG performance.

\subsubsection{Result Comparisons of Out-of-Distribution Generalization on Subsets} 

Table~\ref{tab:sota_ood} lists the out-of-distribution (OOD) generalization performance of $\text{OFS}^{2}\text{Reg}$ and competitive registration methods. On the MICRO-pre2intra subset, $\text{OFS}^{2}\text{Reg}$ achieves the best RMSE, MedE, and P95 of $5.23$, $5.10$, and $14.85$ pixels, respectively. Compared with GDB-ICP, it reduces RMSE by $3.58$ pixels, while GDB-ICP achieves higher mAUC and SR. Compared with EyeLiner, $\text{OFS}^{2}\text{Reg}$ reduces RMSE by $5.36$ pixels and achieves a slightly higher mAUC, whereas EyeLiner obtains a higher SR. Compared with the one-step CGTME, $\text{OFS}^{2}\text{Reg}$ reduces RMSE by $4.02$ pixels and achieves better performance across all reported metrics.

\begin{figure}
    \centering
    \includegraphics[width=0.98\columnwidth]{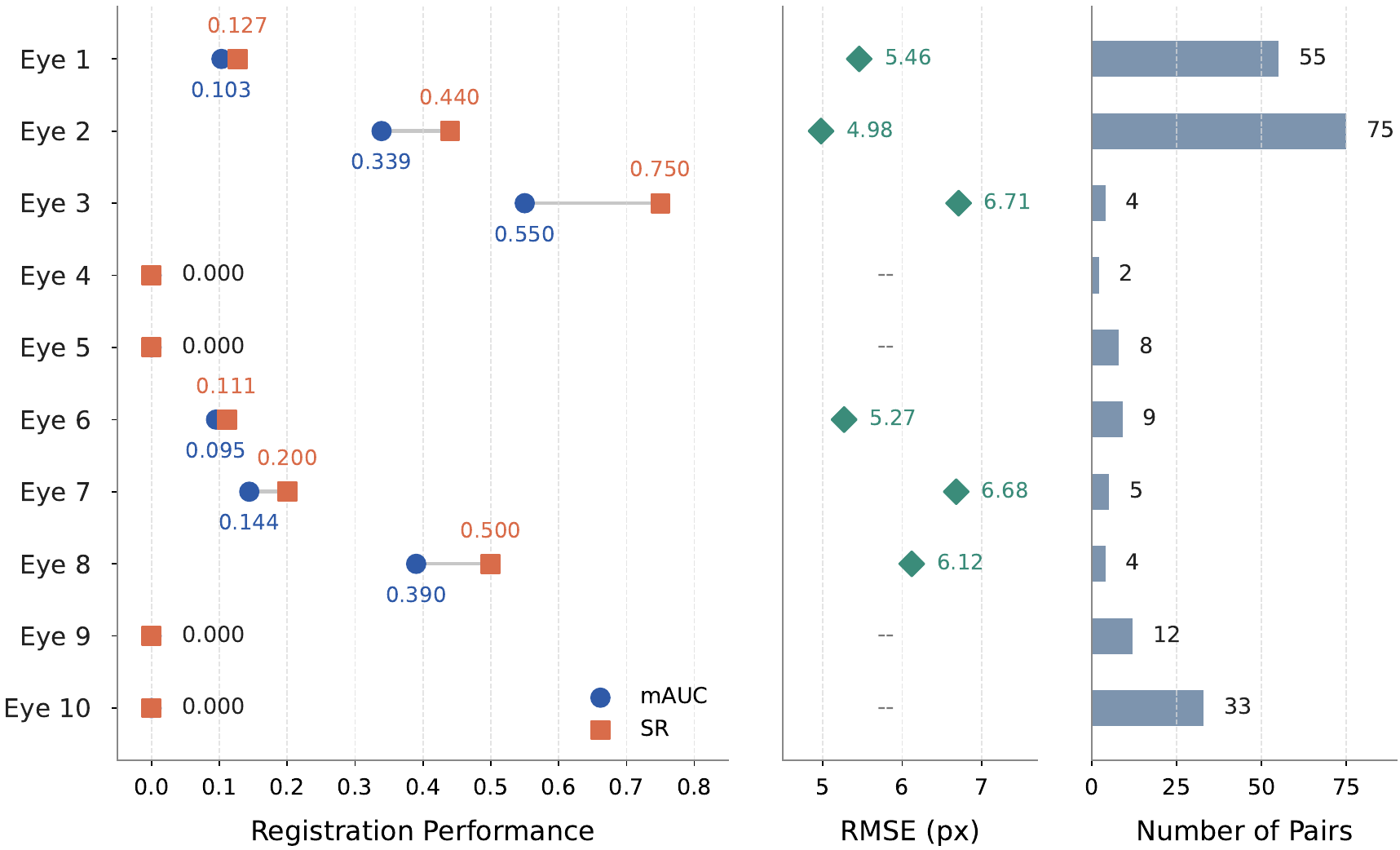}
    \caption{Eye-specific zero-shot registration performance of $\text{OFS}^{2}\text{Reg}$ on MICRO-pre2intra.}
    \label{fig:pre2intra_eye}
\end{figure}

For the following disease-specific, surgical-type, eye-specific, FOV overlap, and failure-case analyses, we use the results from a fixed evaluation run for detailed subgroup and case-level analysis.
To further examine the registration performance under different pathological conditions, Table~\ref{tab:pre2intra_disease} reports the disease-specific results on MICRO-pre2intra. $\text{OFS}^{2}\text{Reg}$ achieves the best performance on choroidal vascular disease, with an RMSE of $4.98$ pixels, an mAUC of $0.339$, and an SR of $0.440$. These results indicate that retinal detachment-related diseases and vitreoretinal interface disorders present greater challenges for registration.
To examine the registration performance under different surgical conditions, Table~\ref{tab:pre2intra_surgery} reports the surgical-type-specific results on MICRO-pre2intra across four surgical types: vitrectomy, PVD induction, ILM staining, and membrane peeling. It sees that $\text{OFS}^{2}\text{Reg}$ achieves the best performance on vitrectomy, with an RMSE of $5.12$ pixels, an mAUC of $0.243$, and an SR of $0.313$. These results indicate that ILM staining and membrane peeling present greater challenges for registration.
Figure~\ref{fig:pre2intra_eye} reports the eye-specific performance of $\text{OFS}^{2}\text{Reg}$ on MICRO-pre2intra, showing performance variation across the 10 eyes together with the corresponding number of image pairs for each eye. Table~\ref{tab:pre2intra_fov_overlap} analyzes registration performance across different FOV overlap ranges. As the FOV overlap increases, registration performance improves substantially: for pairs with FOV overlap below 20\%, the mAUC and SR are only $0.009$ and $0.012$, whereas for pairs with FOV overlap of at least 50\%, they increase to $0.699$ and $0.909$, respectively. These results indicate that limited FOV overlap is an important challenge for pre-to-intraoperative retinal image registration.

Similarly, on the MICRO-pre2post subset, $\text{OFS}^{2}\text{Reg}$ achieves the best RMSE and P95 of $4.10$ and $11.60$ pixels, respectively, together with the highest mAUC of $0.632$ and SR of $0.747$. Compared with GDB-ICP and VDD-Reg, $\text{OFS}^{2}\text{Reg}$ reduces RMSE by $0.98$ and $3.48$ pixels, respectively. Compared with SuperJunction and CGTME, $\text{OFS}^{2}\text{Reg}$ improves mAUC by $0.129$ and $0.060$, and SR by $0.140$ and $0.060$, respectively. All in all, these results manifest the OOD generalization of $\text{OFS}^{2}\text{Reg}$ across different ophthalmic surgical phases.

\begin{table}[t]
\centering
\caption{Disease-specific registration performance of $\text{OFS}^{2}\text{Reg}$ on the MICRO-pre2intra subset.}
\label{tab:pre2intra_disease}
\renewcommand{\arraystretch}{1}
\setlength{\tabcolsep}{4pt}

\resizebox{0.98\columnwidth}{!}{%
\begin{tabular}{lcccc}
\toprule
\textbf{Disease Category} &
\textbf{Pairs} &
\textbf{RMSE $\downarrow$} &
\textbf{mAUC $\uparrow$} &
\textbf{SR $\uparrow$} \\
\midrule

Retinal detachment-related diseases
& 69
& 5.61
& 0.105
& 0.130 \\

Choroidal vascular disease
& 75
& 4.98
& 0.339
& 0.440 \\

Vitreoretinal interface disorders
& 54
& 6.70
& 0.054
& 0.074 \\

Other
& 9
& 5.27
& 0.095
& 0.111 \\

\midrule
Overall
& 207
& 5.28
& 0.176
& 0.227 \\

\bottomrule
\end{tabular}%
}
\end{table}

\begin{table}[t]
\centering
\caption{Zero-shot registration performance of $\text{OFS}^{2}\text{Reg}$ across different surgical types in MICRO-pre2intra.}
\label{tab:pre2intra_surgery}
\renewcommand{\arraystretch}{1} 
\begin{tabular}{lcccc}
\toprule
\textbf{Surgical Type} &
\textbf{Pairs} &
\textbf{RMSE $\downarrow$} &
\textbf{mAUC $\uparrow$} &
\textbf{SR $\uparrow$} \\
\midrule
Vitrectomy       & 134 & 5.12 & 0.243 & 0.313 \\
PVD induction    & 10  & 5.33 & 0.093 & 0.100 \\
ILM staining     & 14  & 6.68 & 0.051 & 0.071 \\
Membrane peeling & 49  & 6.71 & 0.045 & 0.061 \\
\midrule
Overall          & 207 & 5.28 & 0.176 & 0.227 \\
\bottomrule
\end{tabular}
\end{table}

\begin{table}[t]
\centering
\caption{Zero-shot registration performance of
$\text{OFS}^{2}\text{Reg}$ across different FOV overlap ranges
in MICRO-pre2intra.}
\label{tab:pre2intra_fov_overlap}
\renewcommand{\arraystretch}{1}
\setlength{\tabcolsep}{4pt}

\resizebox{0.98\columnwidth}{!}{%
\begin{tabular}{lcccc}
\toprule
\textbf{FOV Overlap} &
\textbf{Pairs} &
\textbf{RMSE $\downarrow$} &
\textbf{mAUC $\uparrow$} &
\textbf{SR $\uparrow$} \\
\midrule
$<20\%$ &
83 &
7.32 &
0.009 &
0.012 \\

$20\%\leq\mathrm{FOV}<50\%$ &
80 &
5.41 &
0.063 &
0.075 \\

$\geq50\%$ &
44 &
5.20 &
0.699 &
0.909 \\

\midrule
Overall &
207 &
5.28 &
0.176 &
0.227 \\

\bottomrule
\end{tabular}%
}
\end{table}

\begin{figure}
    \centering
    \includegraphics[width=0.9\linewidth]{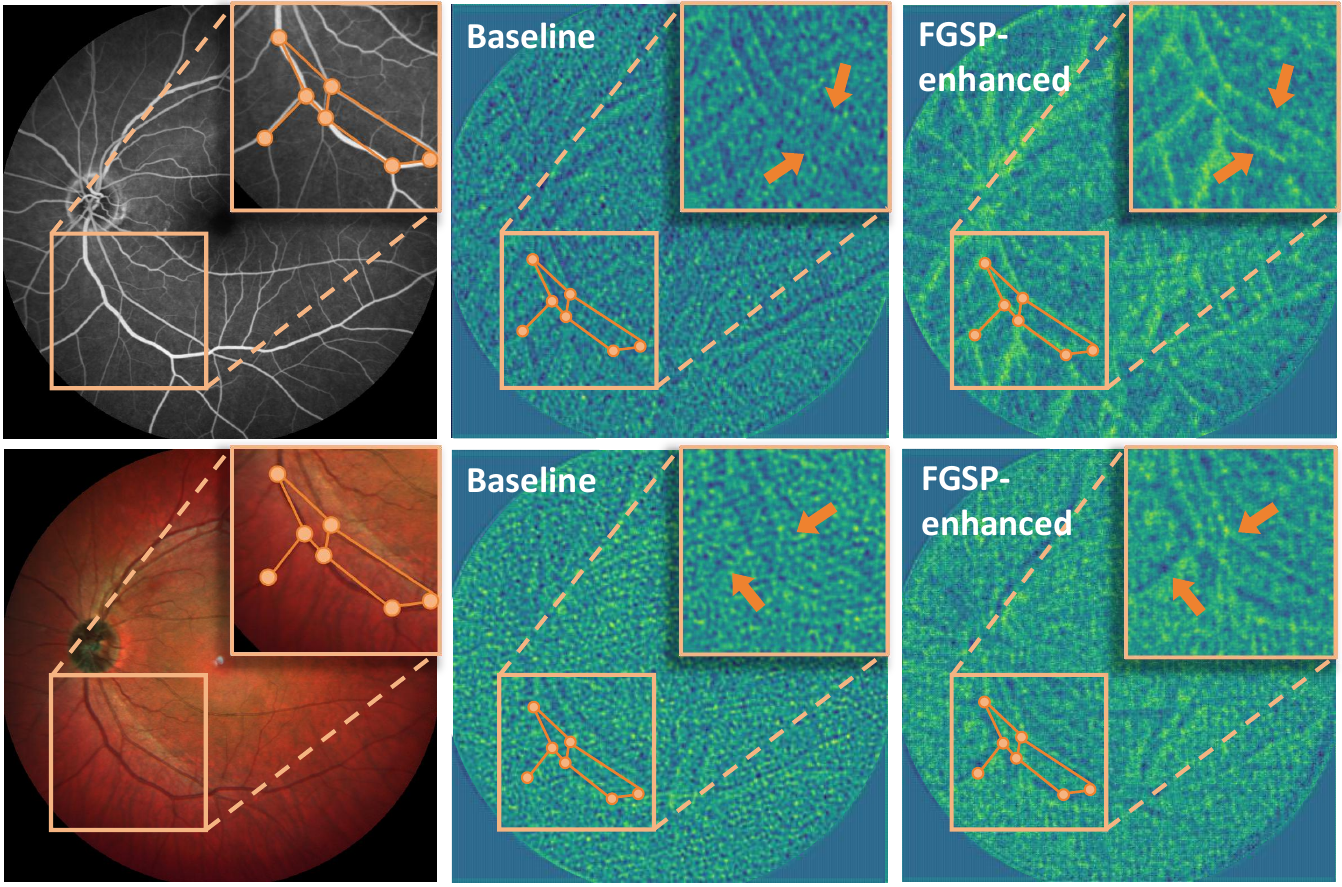}
    \caption{Feature descriptor visualizations of the baseline and the baseline with FGSP.}
    \label{fig:descriptor_viz}
\end{figure}

\subsubsection{External Validation on the CF-FA Dataset}
The CF-FA dataset is constructed from the retinal images provided by Hajeb~\textit{et al.}~\cite{hajeb2012diabetic}. We further annotate corresponding control points following the same annotation protocol as MICRO to establish ground-truth rigid transformations for registration evaluation.
Table~\ref{tab:sota_cffa} reports the external validation results of $\text{OFS}^{2}\text{Reg}$ and competitive registration methods on the CF-FA dataset. $\text{OFS}^{2}\text{Reg}$ achieves the best overall performance, with an RMSE of 3.34 pixels, an mAUC of 0.866, and an SR of 1.000. Compared with the two-step EyeLiner method, $\text{OFS}^{2}\text{Reg}$ reduces RMSE by 0.19 pixels while improving mAUC and SR by 0.052 and 0.085, respectively. Compared with the one-step CGTME, $\text{OFS}^{2}\text{Reg}$ reduces RMSE by 0.16 pixels and improves mAUC and SR by 0.031 and 0.085, respectively. These results demonstrate that $\text{OFS}^{2}\text{Reg}$ maintains strong registration accuracy and robustness on an independent multimodal retinal image dataset, further supporting its generalization beyond the MICRO benchmark.

\begin{table}[t]
\centering
\caption{MRIRG performance comparisons of $\text{OFS}^{2}\text{Reg}$ and state-of-the-art methods on the CF-FA dataset.}
\label{tab:sota_cffa}
\renewcommand{\arraystretch}{1}
\setlength{\tabcolsep}{5pt}
\resizebox{0.98\columnwidth}{!}{%
\begin{tabular}{l l ccc}
    \toprule
    \multirow{2}{*}{\textbf{Category}} &
    \multirow{2}{*}{\textbf{Method}} &
    \multicolumn{3}{c}{\textbf{Metrics}} \\
    \cmidrule(l){3-5}
    & &
    RMSE $\downarrow$ &
    mAUC $\uparrow$ &
    SR $\uparrow$ \\
    \midrule
    
    \multirow{2}{*}{\makecell[l]{Traditional \\ registration}}
    & SIFT~\cite{lowe2004distinctive} (2004)
    & 17.30 & 0.007 & 0.034 \\
    
    & GDB-ICP~\cite{yang2007registration} (2007)
    & 4.34 & 0.215 & 0.254 \\
    
    \midrule
    
    \multirow{15}{*}{\makecell[l]{Deep \\ registration}}
    & \multicolumn{4}{c}{\textit{\textbf{Two-step:}}} \\
    
    & EyeLiner~\cite{veturi2024eyeliner} (2024)
    & 3.53 & 0.814 & 0.915 \\
    
    & VDD-Reg~\cite{wang2024memo} (2024)
    & 4.32 & 0.741 & 0.864 \\
    
    \cmidrule(lr){2-5}
    
    & \multicolumn{4}{c}{\textit{\textbf{One-step:}}} \\
    
    & SuperPoint~\cite{detone2018superpoint} (2018)
    & 3.92 & 0.766 & 0.864 \\
    
    & GLAMpoints~\cite{truong2019glampoints} (2019)
    & 9.08 & 0.060 & 0.085 \\
    
    & R2D2~\cite{revaud2019r2d2} (2019)
    & 11.56 & 0.045 & 0.068 \\
    
    & SuperRetina~\cite{liu2022semi} (2022)
    & 7.80 & 0.034 & 0.051 \\
    
    & SwinU-SuperRetina~\cite{nasser2024reverse} (2024)
    & 5.61 & 0.441 & 0.559 \\
    
    & LK-SuperRetina~\cite{nasser2024reverse} (2024)
    & 4.27 & 0.083 & 0.102 \\
    
    & SuperJunction~\cite{wang2024superjunction} (2024)
    & -- & 0.000 & 0.000 \\
    
    & XFeat~\cite{potje2024xfeat} (2024)
    & 4.94 & 0.774 & \underline{0.996} \\
    
    & XFeat*~\cite{potje2024xfeat} (2024)
    & 4.92 & 0.764 & 0.949 \\
    
    & CGTME~\cite{yan2025two} (2025)
    & \underline{3.50}
    & \underline{0.835}
    & 0.915 \\
    
    & \textbf{$\text{OFS}^{2}\text{Reg}$} (Ours)
    & \textbf{3.34}
    & \textbf{0.866}
    & \textbf{1.000} \\
    
    \bottomrule
\end{tabular}%
}
\end{table}

\subsection{Generalization and Visualization Analysis}

\begin{table}
    \centering
    \caption{Generalization validation of  $\text{OFS}^{2}\text{Reg}$ across classical CNNs and recent foundation models. Results are obtained from a single run for each backbone configuration.}
    \label{tab:backbone_generalization}
    \renewcommand{\arraystretch}{1} 
    \resizebox{0.98\columnwidth}{!}{%
    \begin{tabular}{llccc} 
        \toprule
        \multirow{2}{*}{Backbone} & \multirow{2}{*}{Method} & \multicolumn{3}{c}{Metrics} \\
        \cmidrule(l){3-5}
        & & RMSE $\downarrow$ & mAUC $\uparrow$ & SR $\uparrow$ \\
        \midrule
        
        \multicolumn{5}{c}{\textit{\textbf{Classic CNNs:}}} \\
        \quad VGG-16~\cite{Simonyan2014VeryDC} 
        & Baseline & 8.46 & 0.150 & 0.216 \\ 
        & \textbf{+ $\text{OFS}^{2}\text{Reg}$} & \textbf{8.33} & \textbf{0.240} & \textbf{0.341} \\ 
        \cmidrule(lr){1-5}
        \quad ResNet-50~\cite{he2016deep} 
        & Baseline & 6.33 & 0.333 & 0.425 \\ 
        & \textbf{+ $\text{OFS}^{2}\text{Reg}$} & \textbf{5.08} & \textbf{0.641} & \textbf{0.784}  \\ 
        \midrule

        \multicolumn{5}{c}{\textit{\textbf{Foundation Models:}}} \\
        \quad ViT-B/16~\cite{dosovitskiy2021an} 
        & Baseline & 9.70 & 0.097 & 0.165 \\ 
        & \textbf{+ $\text{OFS}^{2}\text{Reg}$} & \textbf{9.28} & \textbf{0.139} & \textbf{0.219} \\ 
        \cmidrule(lr){1-5}
        \quad CLIP (ViT-L/14)~\cite{radford2021learning} 
        & Baseline & 10.96 & 0.036 & 0.063 \\ 
        & \textbf{+ $\text{OFS}^{2}\text{Reg}$} & \textbf{10.67} & \textbf{0.074} & \textbf{0.129} \\ 
        \cmidrule(lr){1-5}
        \quad DINOv2 (ViT-L/14)~\cite{oquab2024dinov} 
        & Baseline & 11.63 & 0.122 & 0.237 \\ 
        & \textbf{+ $\text{OFS}^{2}\text{Reg}$} & \textbf{10.07} & \textbf{0.182} & \textbf{0.320} \\ 
        \bottomrule
    \end{tabular}%
    }
\end{table}

\subsubsection{Generalization Analysis}
Table~\ref{tab:backbone_generalization} presents the MRIRG results of our $\text{OFS}^{2}\text{Reg}$ across classical CNNs and recent foundation models on the MICRO-pre dataset. We observe that $\text{OFS}^{2}\text{Reg}$ consistently boosts the MRIRG performance of five different backbones.
$\text{OFS}^{2}\text{Reg}$ with ResNet-50~\cite{he2016deep} outperforms other backbone configurations, specifically reaching a peak performance of 0.641 and 0.784 in mAUC and SR.
This is mainly because ResNet-50 is skilled at local feature capturing, which is beneficial for capturing fine vessel structures and local geometric details. In contrast, the relatively limited training data may be insufficient for fully adapting patch-based foundation models to subtle retinal vessel structures and local geometric variations. Nevertheless, $\text{OFS}^{2}\text{Reg}$ improves all five backbones, supporting its generalization ability across different feature extraction architectures.

\subsubsection{Failure Case Analysis}

Table~\ref{tab:failure_analysis} summarizes the dominant failure modes of unsuccessful registrations across the three MICRO subsets, where unsuccessful registrations encompass both \textit{failed} and \textit{inaccurate} cases. The main failure cases include low FOV overlap, instrument or lesion occlusion, poor vessel visibility, anatomical or geometric deformation, and large cross-modal appearance differences. 
These factors diminish the availability and reliability of vessel structure information needed for effective cross-modal keypoint matching.

\begin{figure}
    \centering
    \includegraphics[width=\linewidth]{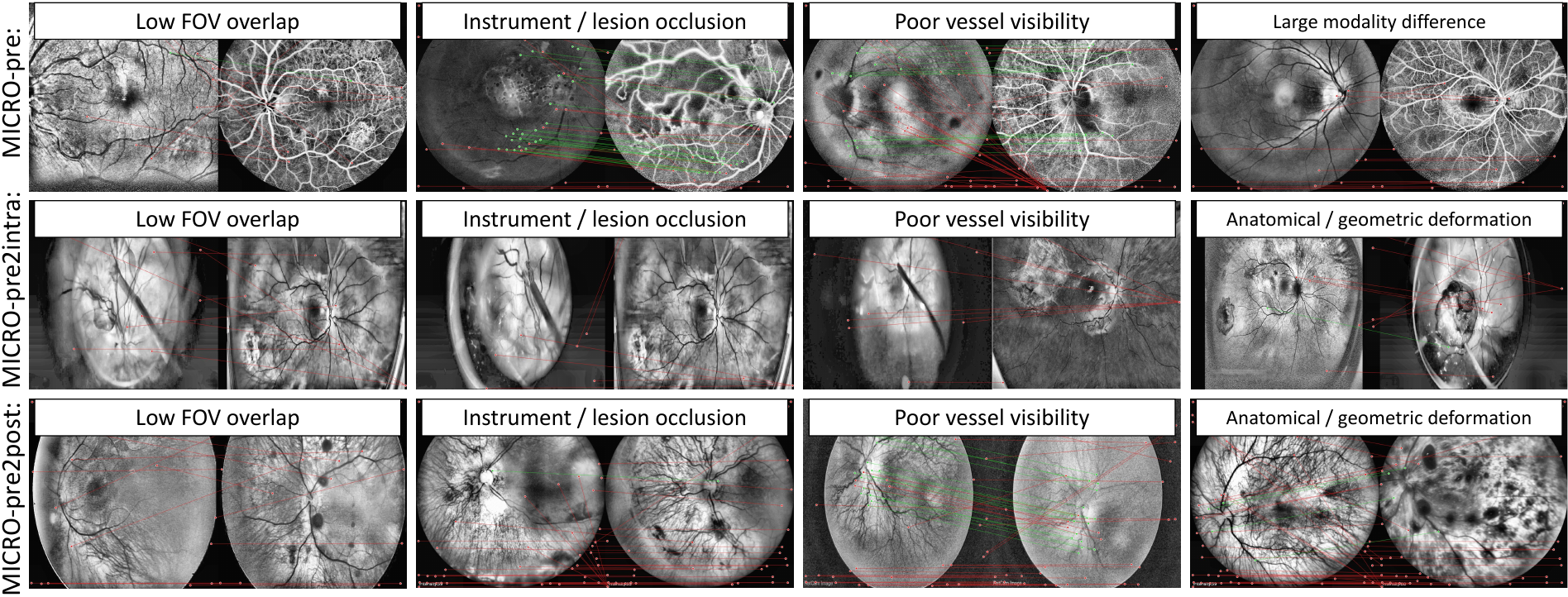}
    \caption{Representative failure cases of $\text{OFS}^{2}\text{Reg}$ on the MICRO dataset. The rows correspond to MICRO-pre, MICRO-pre2intra, and MICRO-pre2post, respectively. Green and red lines indicate correct and incorrect correspondences, respectively.}
    \label{fig:failure_cases}
\end{figure}

On MICRO-pre, low FOV overlap and large cross-modal appearance differences are the main failure modes, accounting for 33.3\% and 26.7\% of unsuccessful cases, respectively. As shown in Fig.~\ref{fig:failure_cases}, large cross-modal appearance differences might lead to erroneous correspondences in invalid background regions. On MICRO-pre2intra, unsuccessful cases are mainly associated with instrument or lesion occlusion (36.7\%), low FOV overlap (30.4\%), and poor vessel visibility (29.7\%), where surgical instruments or lesions can occlude retinal structures and interfere with registration. On MICRO-pre2post, anatomical or geometric deformation is the dominant failure mode (46.2\%), reflecting the influence of postoperative structural changes on local vascular geometry. These observations indicate that a global rigid transformation may be inadequate in situations where significant local deformation occurs.

\begin{table}[t]
\centering
\caption{Failure-mode analysis of unsuccessful registrations of $\text{OFS}^{2}\text{Reg}$ on the MICRO dataset.}
\label{tab:failure_analysis}
\renewcommand{\arraystretch}{1}
\setlength{\tabcolsep}{4pt}
\resizebox{\columnwidth}{!}{%
\begin{tabular}{lccc}
\toprule
\textbf{Failure Mode} &
\textbf{MICRO-pre} &
\textbf{MICRO-pre2intra} &
\textbf{MICRO-pre2post} \\
\midrule
Low FOV overlap
& 20 (33.3\%) & 48 (30.4\%) & 3 (23.1\%) \\

Instrument / lesion occlusion
& 11 (18.3\%) & 58 (36.7\%) & 2 (15.4\%) \\

Poor vessel visibility
& 13 (21.7\%) & 47 (29.7\%) & 2 (15.4\%) \\

Anatomical / geometric deformation
& 0 (0.0\%) & 5 (3.2\%) & 6 (46.2\%) \\

Large modality difference
& 16 (26.7\%) & 0 (0.0\%) & 0 (0.0\%) \\
\midrule
Total unsuccessful cases
& 60 & 158 & 13 \\
\bottomrule
\end{tabular}%
}
\end{table}

\subsubsection{Feature Descriptors Visualization}
Fig.~\ref{fig:descriptor_viz} presents the visualizations of feature descriptors generated by FGSP and the baseline based on representative multimodal retinal image pairs. The visual results show that our FGSP-enhanced feature descriptors effectively accentuate more prominent and complete vessel structures compared to the baseline, especially vessel structures with ambiguous patterns and high noise. Therefore, Fig.~\ref{fig:descriptor_viz} proves that FGSP is beneficial to vessel structure topology constraints construction in feature descriptors by emphasizing significant vessel structures.

\subsubsection{Qualitative Registration Results Visualization}
\begin{figure*}
    \centering
    \includegraphics[width=0.97\linewidth]{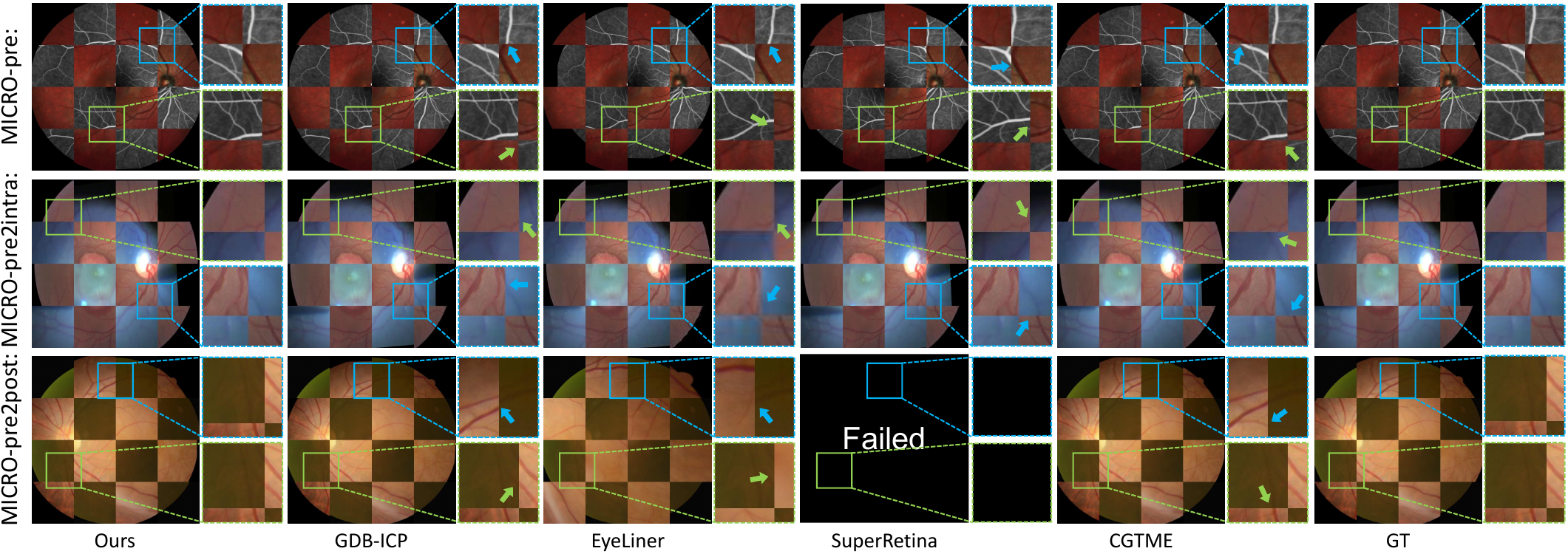}
    \caption{Qualitative MRIRG comparisons of $\text{OFS}^{2}\text{Reg}$, GDB-ICP, EyeLiner, SuperRetina, and CGTME, by selecting cross-modal retinal images across three ophthalmic surgery phases (e.g., preoperative, intraoperative, and postoperative on the MICRO benchmark). }
    \label{fig:contrast}
\end{figure*}

Fig.~\ref{fig:contrast} shows the qualitative registration results of $\text{OFS}^{2}\text{Reg}$, GDB-ICP, EyeLiner, SuperRetina, and CGTME in mosaic format. Here, we select cross-modal retinal image pairs from three different ophthalmic surgery phases in the MICRO subsets, along with the ground-truth (GT) alignment for reference. It can be observed that $\text{OFS}^{2}\text{Reg}$ effectively registers both main vessel trunks and fine vessels with higher precision than other competitive methods. In contrast, other comparable methods exhibit noticeable vessel misalignments and apparent discontinuities at the mosaic boundaries, indicating inaccurate rigid transformation. The visual results show that our $\text{OFS}^{2}\text{Reg}$ obtains a visual appearance that most closely resembles the GT alignment, supporting our motivation to improve the accuracy of pixel-level cross-modal vessel structure alignment from the perspective of keypoint topology constraint construction.

\section{Discussion and Limitations}

In this paper, we construct the MICRO benchmark for multimodal retinal image rigid registration across different surgical phases and develop $\text{OFS}^{2}\text{Reg}$ for robust cross-modal retinal image registration. Experimental results demonstrate the effectiveness of our method on MICRO and an external CF-FA dataset. However, there are still some limitations in the field of MRIRG:

\begin{itemize}
    \item MICRO-pre2intra and MICRO-pre2post contain only 10 and 50 eyes, respectively. Therefore, we plan to collect larger and more diverse surgical cohorts with different eye conditions for further validation. Moreover, instrument or lesion occlusion, low FOV overlap, and poor vessel visibility also remain major challenges for intraoperative registration,  limiting the potential improvements in the performance of deep registration methods.
    
    \item Our method mainly focuses on rigid registration, while intraoperative and postoperative retinal images may contain local non-rigid deformation that cannot be fully corrected by a global rigid transformation.
    
    \item The cross-modal training stage relies on ground-truth rigid transformations derived from manually annotated control-point correspondences, which restricts direct adaptation to new modality pairs without similar control-point annotations.
\end{itemize}

To address these limitations, we plan further to expand MICRO with larger, more diverse surgical cohorts and to develop more robust cross-modal keypoint-matching strategies to tackle challenging conditions, including instrument or lesion occlusion, low FOV overlap, and poor vessel visibility. We also plan to extend our proposed $\text{OFS}^{2}\text{Reg}$ from rigid registration to rigid-to-deformable registration, in which rigid registration corrects large global displacements and deformable registration further refines local anatomical alignment. In addition, we will modify training strategies to reduce the dependence on manually annotated geometric supervision, thereby improving the adaptability of the proposed framework to new retinal modality combinations.

\section{Conclusion}
In this paper, we introduce the publicly accessible MICRO dataset to advance the MRIRG techniques. We rethink the MRIRG task from the perspective of keypoint topology constraints, providing an alternative means for one-step registration technique design. Under this perspective, we propose a novel one-step $\text{OFS}^{2}\text{Reg}$ framework for MRIRG by employing a well-designed  FGSP descriptor and a novel $\text{CMVS}^{2}$ learning method to construct keypoint topology constraints through exploiting cross-modal vessel structure similarities. 
The extensive experiments on the MICRO dataset demonstrate that $\text{OFS}^{2}\text{Reg}$ generally performs better than state-of-the-art methods across different surgical-phase subsets, particularly in registration accuracy.
In the future, we plan to provide more theoretical analysis of MRIRG to further investigate its generalization and robustness across other cross-modal image rigid registration tasks.

\section*{Declaration of Competing Interest
}
None.

\section*{Acknowledgments}
This work was supported in part by 
National Key R\&D Program of China (Grant No.2024YFC2510800), National Natural Science Foundation of China (Grant No.62602659), and Shenzhen Medical Research Fund (Grant No.D2402014).

\bibliographystyle{cas-model2-names}

\bibliography{references}


\end{document}